\documentclass{article} 
\usepackage{iclr2019_conference,times}

\usepackage{amsmath,amsfonts,bm,multirow}

\def\eqref#1{equation~\ref{#1}}
\def\Eqref#1{Equation~\ref{#1}}

\def\1{\bm{1}}

\DeclareMathAlphabet{\mathsfit}{\encodingdefault}{\sfdefault}{m}{sl}
\SetMathAlphabet{\mathsfit}{bold}{\encodingdefault}{\sfdefault}{bx}{n}

\newcommand{\E}{\mathbb{E}}

\newcommand{\R}{\mathbb{R}}

\DeclareMathOperator{\Tr}{Tr}

\DeclareMathOperator{\IS}{IS} 
\DeclareMathOperator{\FID}{FID} 

\newcommand{\celebahq}{\textsc{celeba-hq}}
\newcommand{\lsun}{\textsc{lsun-bedroom}}
\newcommand{\cifar}{\textsc{cifar10}}
\newcommand{\imagenet}{\textsc{imagenet}}

\usepackage{multirow}
\usepackage{hyperref}
\usepackage{xcolor}
\hypersetup{
    colorlinks,
    anchorcolor={blue!50!black},
    linkcolor={blue!50!black},
    citecolor={blue!50!black},
    urlcolor={blue}
}
\usepackage{url}
\usepackage{epsfig}
\usepackage{booktabs}
\usepackage{subcaption}
\usepackage{comment}
\usepackage{array}

\title{On Self Modulation for Generative Adversarial Networks}

\author{Ting Chen\thanks{Work done at Google.}  \\
University of California, Los Angeles \\
\texttt{tingchen@cs.ucla.edu}
\And
Mario Lucic, Neil Houlsby, Sylvain Gelly\\
Google Brain \\
\texttt{\{lucic,neilhoulsby,sylvaingelly\}@google.com}
}

\newcommand{\nop}[1]{}

\iclrfinalcopy 
\begin{document}

\maketitle

\begin{abstract}
Training Generative Adversarial Networks (GANs) is notoriously challenging.
We propose and study an architectural modification, \emph{self-modulation}, which improves GAN performance across different data sets, architectures, losses, regularizers, and hyperparameter settings.
Intuitively, self-modulation allows the intermediate feature maps of a generator to change as a function of the input noise vector.
While reminiscent of other conditioning techniques, it requires no labeled data.
In a large-scale empirical study we observe a relative decrease of $5\%-35\%$ in \textsc{fid}.
Furthermore, all else being equal, adding this modification to the generator leads to improved performance in $124/144$ ($86\%$) of the studied settings.
Self-modulation is a simple architectural change that requires no additional parameter tuning, which suggests that it can be applied readily to any GAN.\footnote{Code at \url{https://github.com/google/compare_gan}}
\end{abstract}

\section{Introduction}

Generative Adversarial Networks (GANs) are a powerful class of
generative models successfully applied to a variety of tasks such as
image
generation~\citep{zhang2017stackgan,miyato2018spectral,karras2017progressive},
learned compression~\citep{tschannen2018distributionpreserving},
super-resolution~\citep{ledig2017photo},
inpainting~\citep{pathak2016context}, and domain
transfer~\citep{pix2pix2016,zhu2017unpaired}.

Training GANs is a notoriously challenging
task~\citep{goodfellow2014generative,arjovsky2017wasserstein,lucic2018}
as one is searching in a high-dimensional parameter space for a Nash equilibrium of a non-convex game.
As a practical remedy one applies
(usually a variant of) stochastic gradient descent, which can be
unstable and lack guarantees~\cite{salimans2016improved}. As a result, one of the main
research challenges is to stabilize GAN training. Several approaches
have been proposed, including varying the underlying divergence between the
model and data
distributions~\citep{arjovsky2017wasserstein,mao2016least},
regularization and normalization
schemes~\citep{gulrajani2017improved,miyato2018spectral}, optimization
schedules ~\citep{karras2017progressive}, and specific neural
architectures~\citep{radford2016,zhang2018self}. A particularly
successful approach is based on conditional generation; where
the generator (and possibly discriminator) are given side information, for
example class labels~\cite{mirza2014conditional,odena2017,miyato2018cgans}. In
fact, state-of-the-art conditional GANs inject side information via
conditional batch normalization (CBN)
layers~\citep{de2017modulating,miyato2018cgans,zhang2018self}. While
this approach does help, a major drawback is that it requires external
information, such as labels or embeddings, which is not always available.

In this work we show that GANs benefit from \emph{self-modulation}
layers in the generator.
Our approach is motivated by Feature-wise Linear Modulation in supervised
learning~\citep{perez2018,de2017modulating}, with one key difference:
instead of conditioning on external information, we condition on the
generator's own input.
As self-modulation requires a simple change which is easily applicable to all popular generator architectures,
we believe that is a useful addition to the GAN toolbox.

\textbf{Summary of contributions.}
We provide a simple yet effective technique that can added universally to yield better GANs.
We demonstrate empirically that for a wide variety of
settings (loss functions, regularizers and normalizers, neural
architectures, and optimization settings) that the proposed approach yields
between a $5\%$ and $35\%$ improvement in sample quality.
When using fixed hyperparameters settings our approach
outperforms the baseline in $86\% (124/144)$ of cases.
Further, we show that self-modulation still helps even if label information is available.
Finally, we discuss the effects of this method in light of recently proposed diagnostic tools,
generator conditioning~\citep{odena2018generator}
and precision/recall for generative models~\citep{sajjadi2018assessing}.

\section{Self-Modulation for Generative Adversarial Networks}\label{sec:model}

\begin{figure}[t]
\begin{center}
\includegraphics[height=3.3cm, trim={0 0 0 0},clip]{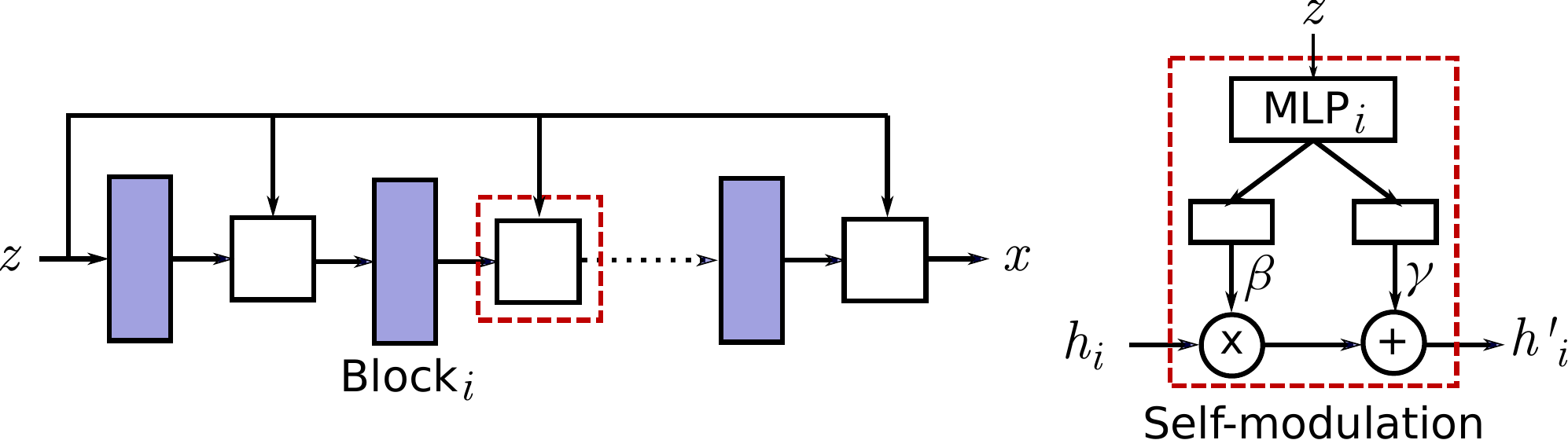}
\end{center}
\caption{\label{fig:simple_abn1}
  (a) The proposed Self-Modulation framework for a generator network,
  where middle layers are directly modulated as a function of the
  generator input $\bm z$. (b) A simple MLP based modulation function
  that transforms input $\bm z$ to the modulation variables $\bm
  \beta(\bm z)$ and $\bm \gamma(\bm z)$.}
\end{figure}

Several recent works observe that conditioning the generative
process on side information (such as labels or class embeddings) leads
to improved
models~\citep{mirza2014conditional,odena2017,miyato2018cgans}. Two
major approaches to conditioning on side information $\bm s$ have
emerged: (1) Directly concatenate the side information $\bm s$ with
the noise vector $\bm z$~\citep{mirza2014conditional}, i.e. $\bm z' =
[\bm s, \bm z]$. (2) Condition the hidden layers directly on $\bm s$,
which is usually instantiated via conditional batch
normalization~\citep{de2017modulating,miyato2018cgans}.

Despite the success of conditional approaches, two concerns arise.
The first is practical; side information is often unavailable.  The
second is conceptual; unsupervised models, such as GANs, seek to model
data without labels. Including them side-steps the challenge and value
of unsupervised learning.

We propose \textit{self-modulating} layers for the generator network.
In these layers the hidden activations are modulated as a function of latent vector $\bm z$.
In particular, we apply modulation in a feature-wise fashion which allows the
model to re-weight the feature maps as a function of the input.
This is also motivated by the FiLM layer for supervised models~\citep{perez2018,de2017modulating}
in which a similar mechanism is used to condition a supervised network on side information.

Batch normalization~\citep{ioffe2015batch} can
improve the training of deep neural nets, and it is widely used in
both discriminative and generative modeling~\citep{szegedy2015going,radford2016,miyato2018spectral}.
It is thus present in most modern networks, and provides a convenient entry point for self-modulation.
Therefore, we present our method in the context of its application via batch normalization.
In batch normalization the activations of a layer, $\bm h$, are transformed as
\begin{equation}
\bm h'_\ell = \bm\gamma \odot \frac{\bm h_\ell - \bm \mu}{\bm \sigma} + \bm \beta,
\label{eqn:film}
\end{equation}
where $\bm\mu$ and $\bm\sigma^2$ are the estimated mean and variances
of the features across the data, and $\bm\gamma$ and $\bm\beta$ are
learnable scale and shift parameters.

\textbf{Self-modulation for unconditional (without side information) generation.}
In this case the proposed method replaces the non-adaptive parameters
$\bm\beta$ and $\bm\gamma$ with input-dependent $\bm\beta(\bm z)$ and
$\bm\gamma(\bm z)$, respectively.
These are parametrized by a neural network applied
to the generator's input (Figure~\ref{fig:simple_abn1}).
In particular, for layer $\ell$, we compute
\begin{equation}
\bm h'_{\ell} = \bm \gamma_{\ell}(\bm z) \odot \frac{\bm h_{\ell} - \bm \mu}{\bm \sigma} + \bm\beta_{\ell}(\bm z)\,
\end{equation}
In general, it suffices that $\bm\gamma_\ell(\cdot)$ and
$\bm\beta_\ell(\cdot)$ are differentiable. In this work, we use a
small one-hidden layer feed-forward network (MLP) with ReLU activation
applied to the generator input $\bm z$. Specifically, given parameter matrices
$U^{(\ell)}$ and $V^{(\ell)}$, and a bias vector $\bm b^{(\ell)}$, we compute
\[
\bm \gamma_\ell(\bm z) = V^{(\ell)} \max(0, U^{(\ell)}\bm z + \bm b^{(\ell)}).
\]
We do the same for $\beta(\bm z)$ with independent parameters.

\textbf{Self-modulation for conditional (with side information) generation.}
Having access to side information proved to be useful for conditional
generation. The use of labels in the generator (and possibly
discriminator) was introduced by~\citet{mirza2014conditional} and
later adapted by \citet{odena2017,miyato2018cgans}. In case that side
information is available (e.g.\,class labels $y$), it can be readily
incorporated into the proposed method. This can be achieved by simply
composing the information $y$ with the input $\bm z \in \R^d$ via some
learnable function $g$, i.e. $\bm z' = g(y, \bm z)$. In this work we
opt for the simplest option and instantiate $g$ as a bi-linear
interaction between $\bm z$ and two trainable embedding functions
$E,E': Y \rightarrow \R^d$ of the class label $y$, as
\begin{equation}
  \label{eq:conditioning}
\bm z' = \bm z + \mathrm E(y) + \bm z \odot \mathrm E'(y).
\end{equation}
This conditionally composed $\bm z'$ can be directly used in
\Eqref{eqn:film}. Despite its simplicity, we demonstrate that it
outperforms the standard conditional models.

\begin{table}[]
\centering
\small
\caption{\label{tab:method_comp} Techniques for generator conditioning and modulation.}
\begin{tabular}{lcc}
  \toprule
  & \textbf{Only first layer} & \textbf{Other Arbitrary layers}  \\ \midrule
  \shortstack[l]{\textbf{Side information} $\bm s$\\ \hfill}  & \shortstack[c]{N/A \\ \hfill} & \shortstack[c]{Conditional batch normalization\\ \citep{de2017modulating,miyato2018cgans}}     \\ \midrule
\shortstack[l]{\textbf{Latent vector} $\bm z$ \\ \hfill} & \shortstack[c]{Unconditional Generator\\\citep{goodfellow2014generative}} & \shortstack[c]{(Unconditional) Self-Modulation (this work)\\ \hfill} \\ \midrule
\textbf{Both $\bm s$ and $\bm z$} & \shortstack[c]{Conditional Generator\\\citep{mirza2014conditional}} & (Conditional) Self-Modulation (this work) \\ \bottomrule
\end{tabular}
\end{table}

\textbf{Discussion.} Table \ref{tab:method_comp} summarizes recent
techniques for generator conditioning. While we choose to implement
this approach via batch normalization, it can also operate
independently by removing the normalization part in the
\Eqref{eqn:film}. We made this pragmatic choice due to the fact that
such conditioning is common~\citep{radford2016,miyato2018spectral,miyato2018cgans}.

The second question is whether one benefits from more complex
modulation architectures, such as using an attention
network~\citep{vaswani2017attention} whereby $\beta$ and $\gamma$
could be made dependent on all upstream activations, or constraining
the elements in $\bm\gamma$ to $(0,1)$ which would yield a similar
gating mechanism to an LSTM cell~\citep{hochreiter1997long}. Based on
initial experiments we concluded that this additional complexity does
not yield a substantial increase in performance.

\section{Experiments}
\label{sec:experiments}

We perform a large-scale study of self-modulation to demonstrate that
this method yields robust improvements in a variety of settings.  We
consider loss functions, architectures, discriminator
regularization/normalization strategies, and a variety of
hyperparameter settings collected from recent
studies~\citep{radford2016,gulrajani2017improved,miyato2018spectral,lucic2018,kurach2018}.
We study both unconditional (without labels) and conditional (with
labels) generation.  Finally, we analyze the results through the lens
of the condition number of the generator's Jacobian as suggested
by~\cite{odena2018generator}, and precision and recall as defined
in~\citet{sajjadi2018assessing}.

\subsection{Experimental Settings}\label{sec:settings}

\textbf{Loss functions.}
We consider two loss functions. The first one is the non-saturating loss proposed in~\citet{goodfellow2014generative}:
\begin{align*}
V_D(G, D) &= \mathbb{E}_{\bm x\sim P_d(\bm x)}[\log \sigma(D(\bm x))] + \mathbb{E}_{\bm z\sim P(\bm z)}[\log (1- \sigma(D(G(\bm z))))] \\
V_G(G, D) &=  -\mathbb{E}_{\bm z \sim P(\bm z)}[\log \sigma(D(G(\bm z)))]
\end{align*}
The second one is the hinge loss used in~\citet{miyato2018spectral}:
\begin{align*}
V_D(G, D) &= \mathbb{E}_{\bm x\sim P_d(\bm x)}[\min(0, -1 + D(\bm x))] + \mathbb{E}_{\bm z \sim P(\bm z)}[\min(0, -1 -D(G(\bm z)))] \\
V_G(G, D) &=  -\mathbb{E}_{\bm z \sim P(\bm z)}[D(G(\bm z))]
\end{align*}

\textbf{Controlling the Lipschitz constant of the discriminator.}
The discriminator's Lipschitz constant is a central quantity analyzed in
the GAN literature~\citep{ miyato2018spectral,zhou2018understanding}. We
consider two state-of-the-art techniques: gradient
penalty~\citep{gulrajani2017improved}, and spectral
normalization~\citep{miyato2018spectral}. Without normalization and
regularization the models can perform poorly on some datasets.  For
the gradient penalty regularizer we consider regularization strength $\lambda
\in \{1,10\}$.

\textbf{Network architecture.}
We use two popular architecture types: one based on
DCGAN~\citep{radford2016}, and another from~\citet{miyato2018spectral}
which incorporates residual connections~\citep{he2016}.  The details
can be found in the appendix.

\textbf{Optimization hyper-parameters.}
We train all models for $100$k generator steps with the Adam
optimizer~\citep{kingma2014adam} (We also perform a subset of the
studies with $500$K steps and discuss it in.
We test two popular settings of the
Adam hyperparameters $(\beta_1, \beta_2)$: $(0.5, 0.999)$ and $(0,
0.9)$.  Previous studies find that multiple discriminator steps per
generator step can help the
training~\citep{goodfellow2014generative,salimans2016improved}, thus
we also consider both $1$ and $2$ discriminator steps per generator
step\footnote{We also experimented with 5 steps which didn't
outperform the $2$ step setting.}.  In total, this amounts to three
different sets of hyper-parameters for $(\beta_1, \beta_2,
\text{disc\_iter})$: $(0, 0.9, 1)$, $(0, 0.9, 2)$, $(0.5, 0.999, 1)$.
We fix the learning rate to $0.0002$ as in~\citet{miyato2018spectral}.
All models are trained with batch size of 64 on a single nVidia P100
GPU.  We report the best performing model attained during the training
period; although the results follow the same pattern if the final
model is report.

\textbf{Datasets.}
We consider four datasets: \cifar{}, \celebahq{}, \lsun{}, and
\imagenet{}.  The \lsun{} dataset~\citep{yu15lsun} contains around 3M
images.  We partition the images randomly into a test set containing
30588 images and a train set containing the rest.  \celebahq{}
contains 30k images~\citep{karras2017progressive}.  We use the
$128\times128\times3$ version obtained by running the code provided by
the authors\footnote{Available at
\url{https://github.com/tkarras/progressive\_growing\_of\_gans}.}.  We
use 3000 examples as the test set and the remaining examples as the
training set.  \cifar{} contains 70K images ($32\times32\times3$),
partitioned into 60000 training instances and 10000 testing instances.
Finally, we evaluate our method on \imagenet{}, which contains
$1.3$M training images and $50$K test images.  We re-size the images
to $128\times128\times3$ as done in~\citet{miyato2018cgans}
and~\citet{zhang2018self}.

\textbf{Metrics.}
Quantitative evaluation of generative models remains one of the most
challenging tasks. This is particularly true in the context of
implicit generative models where likelihood cannot be effectively
evaluated. Nevertheless, two quantitative measures have recently emerged:
The Inception Score and the Frechet Inception Distance. While both of
these scores have some drawbacks, they correlate well with scores
assigned by human annotators and are somewhat robust.

Inception Score (IS)~\citep{salimans2016improved} posits that
that the conditional label distribution $p(y|\bm x)$ of samples
containing meaningful objects should have low entropy,
while the marginal label distribution $p(y)$ should have high entropy.
Formally, $\IS(G) = \exp(\E_{x \sim G}[\mathrm{d_{KL}}(p(y|\bm x),
p(y)])$.  The score is computed using an Inception
classifier~\citep{szegedy2015going}.  Drawbacks of applying IS to
model comparison are discussed in~\citet{barratt2018note}.

An alternative score, the Frechet Inception
Distance (FID), requires no labeled data~\citep{heusel2017gans}. The
real and generated samples are first embedded into a feature space
(using a specific layer of InceptionNet). Then, a multivariate
Gaussian is fit each dataset and the distance is computed as $\FID(x,
g) = ||\mu_x -\mu_g||_2^2 + \Tr(\Sigma_x + \Sigma_g -
2(\Sigma_x\Sigma_g)^\frac12)$, where $\mu$ and $\Sigma$ denote the
empirical mean and covariance and subscripts $x$ and $g$ denote the
true and generated data, respectively.  FID was shown to be robust to
various manipulations and sensitive to mode dropping~\citep{heusel2017gans}.

\subsection{Robustness experiments for unconditional generation}\label{sec:experiments_unconditional}
To test robustness, we run a Cartesian product of the parameters in
Section~\ref{sec:settings} which results in 36 settings for each dataset
(2 losses, 2 architectures, 3 hyperparameter settings for spectral normalization, and 6 for gradient penalty).
For each setting we run five random seeds for self-modulation and the baseline
(no self-modulation, just batch normalization).
We compute the median score across random seeds which results in $1440$ trained
models.

\begin{figure}[t]
  \centering
  \begin{subfigure}[b]{0.25\textwidth}
    \includegraphics[width=1\textwidth]{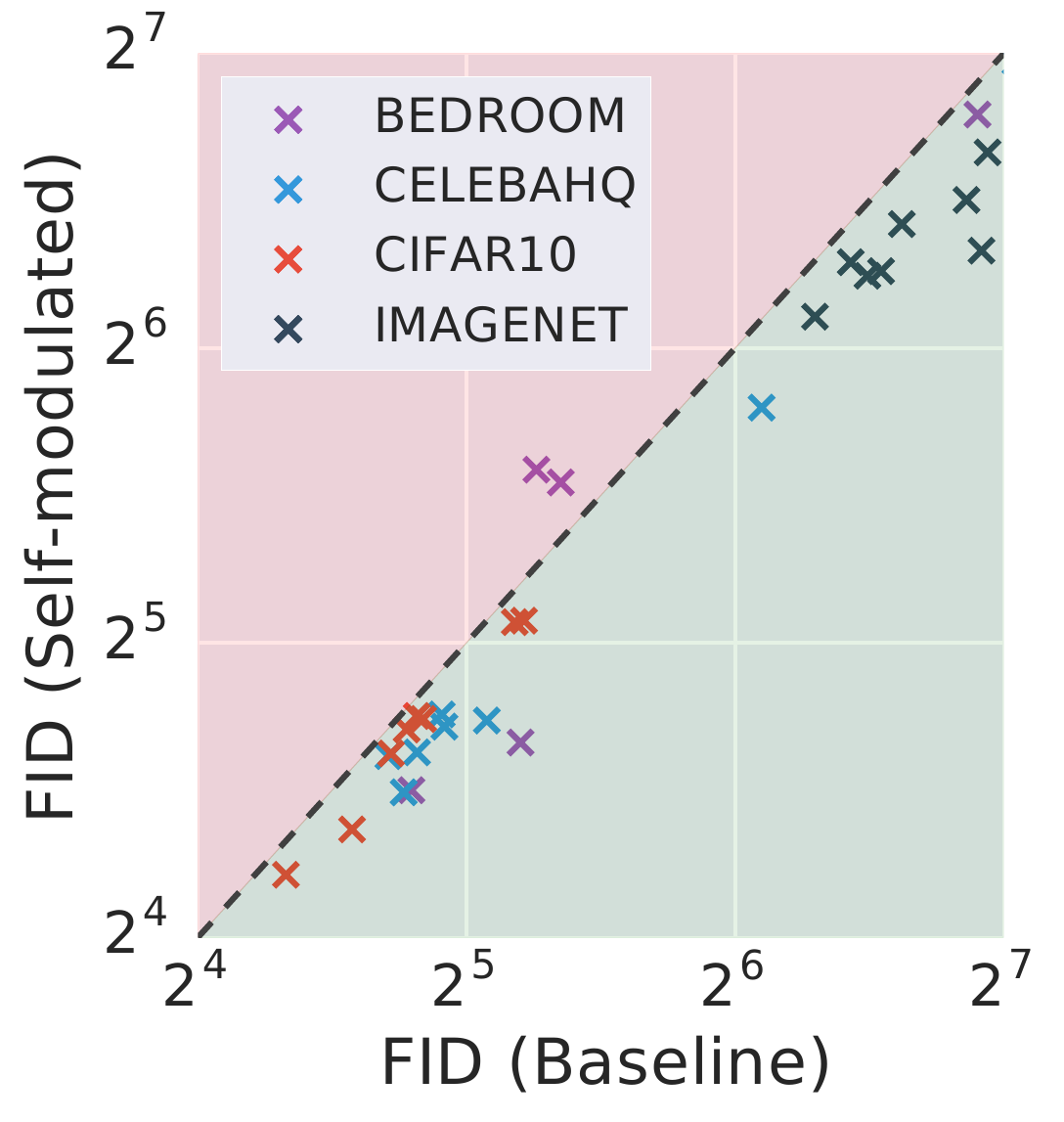}
    \caption{}
  \end{subfigure}\quad
  \begin{subfigure}[b]{0.35\textwidth}
    \includegraphics[width=1\textwidth]{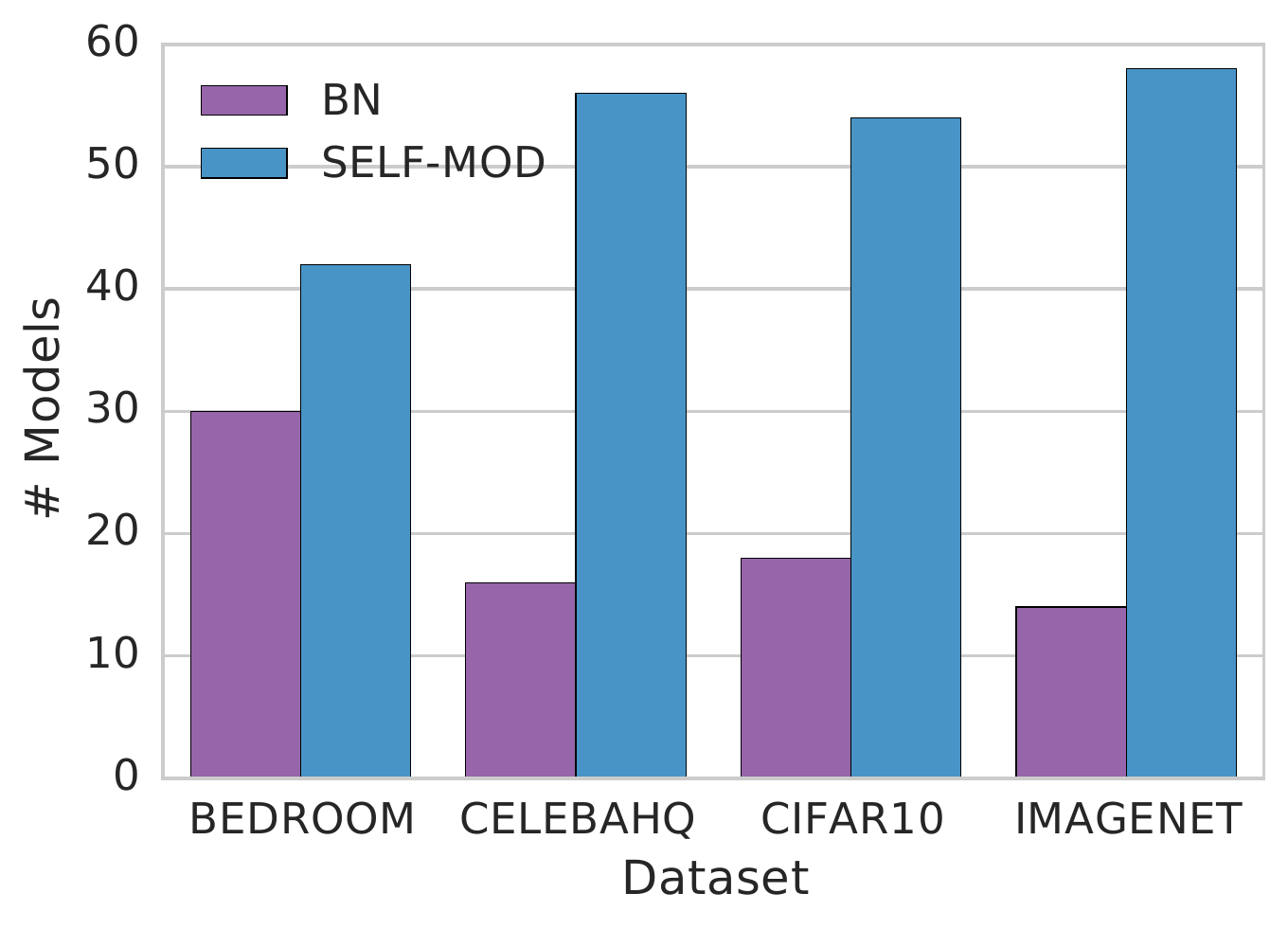}
    \caption{}
  \end{subfigure}\quad
  \begin{subfigure}[b]{0.27\textwidth}
    \includegraphics[width=1\textwidth]{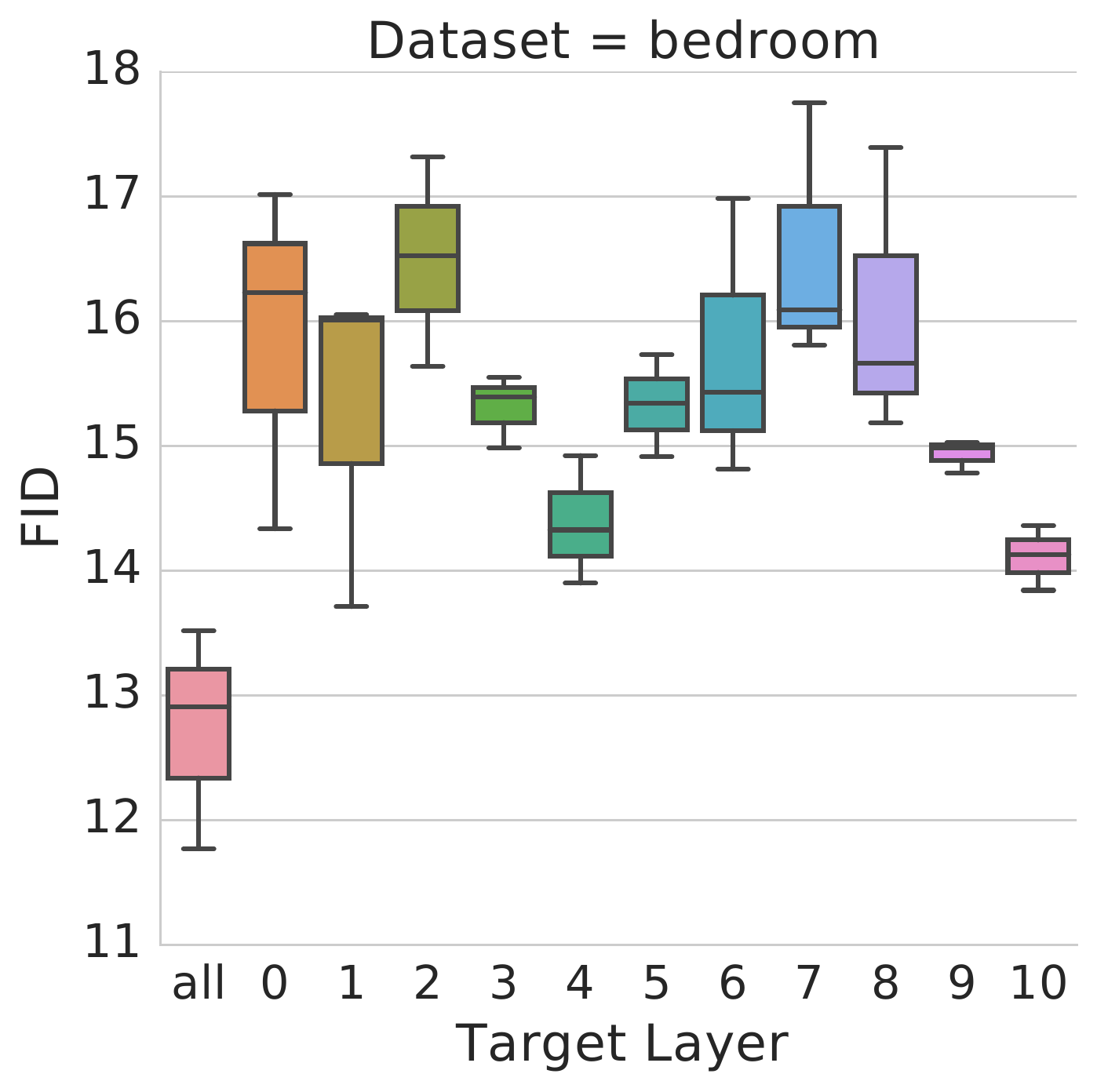}
    \caption{}
  \end{subfigure}
  \caption{\label{fig:fid_main}
    In Figure (a) we observe that the proposed method outperforms the
    baseline in the unpaired setting. Figure (b) shows the number of
    models which fall in 80-th percentile in terms of FID (with reverse
    ordering). We observe that the majority ``good'' models utilize
    self-modulation. Figure (c) shows that applying self-conditioning is
    more beneficial on the later layers, but should be applied to each
    layer for optimal performance. This effect persists across all considered
    datasets, see the appendix.
  }
\end{figure}

We distinguish between two sets of experiments. In the \emph{unpaired
setting} we define the model as the tuple of loss,
regularizer/normalization, neural architecture, and conditioning
(self-modulated or classic batch normalization). For each model
compute the minimum FID across optimization hyperparameters
($\beta_1$, $\beta_2$, $disc\_iters$).
We therefore compare the performance of self-modulation and baseline for each
model after hyperparameter optimization.
The results of this study are reported in Table~\ref{tab:unconditional_raw}, and
the relative improvements are in
Table~\ref{tab:unconditional_reduction} and Figure~\ref{fig:fid_main}.

We observe the following: (1) When using the \textsc{resnet} style
architecture, the proposed method outperforms the baseline \emph{in
all considered settings}.  (2) When using the \textsc{sndcgan}
architecture, it outperforms the baseline in $87.5\%$ of the
cases. The breakdown by datasets is shown in Figure~\ref{fig:fid_main}.  (3)
The improvement can be as high as a $33\%$ reduction in \textsc{fid}.  (4) We
observe similar improvement to the inception score, reported in the
appendix.

In the second setting, the \emph{paired setting}, we assess how
effective is the technique when simply added to an existing model
\emph{with the same set of hyperparameters}. In particular, we fix
everything except the type of conditioning -- the model tuple now
includes the optimization hyperparameters. This results in 36 settings
for each data set for a total of 144 comparisons. We observe that
self-modulation outperforms the baseline in 124/144 settings. These
results suggest that self-modulation can be applied to most GANs even
without additional hyperparameter tuning.

\begin{table}
  \centering
  \caption{\label{tab:unconditional_raw}
    In the unpaired setting (as defined in
    Section~\ref{sec:experiments_unconditional}), we compute the median
    score (across random seeds) and report the best attainable score
    across considered optimization hyperparameters. \textsc{Self-Mod} is
    the method introduced in Section~\ref{sec:model} and \textsc{baseline}
    refers to batch normalization. We observe that the proposed approach
    outperforms the baseline in 30 out of 32 settings. The
    relative improvement is detailed in
    Table~\ref{tab:unconditional_reduction}. The standard error of the median is within $3\%$ in the majority of the settings and is presented in Table~\ref{tab:unconditional_std} for clarity.}
  \small
  {\renewcommand{\arraystretch}{1.2}
  \centering
  \small
  \begin{tabular}{llllrrrr}
    \toprule
    \textsc{Type} & \textsc{Arch} & \textsc{Loss} & \textsc{Method} & \textsc{bedroom} &  \textsc{celebahq} &  \textsc{cifar10} &  \textsc{imagenet}\\
    \midrule
    \multirow{8}{*}{\shortstack{\textsc{Gradient}\\\textsc{penalty}}} & \multirow{4}{*}{\textsc{res}} & \multirow{2}{*}{\textsc{hinge}} & \textsc{self-mod} &    22.62 &     27.03 &    26.93 &     78.31 \\
                  &        &       & \textsc{baseline} &    27.75 &     30.02 &    28.14 &     86.23 \\
    \cline{3-8}
                  &        & \multirow{2}{*}{\textsc{ns}} & \textsc{self-mod} &    25.30 &     26.65 &    26.74 &     85.67 \\
                  &        &       & \textsc{baseline} &    36.79 &     33.72 &    28.61 &     98.38 \\
    \cline{2-8}
                  & \multirow{4}{*}{\textsc{sndc}} & \multirow{2}{*}{\textsc{hinge}} & \textsc{self-mod} &   110.86 &     55.63 &    33.58 &     90.67 \\
                  &        &       & \textsc{baseline} &   119.59 &     68.51 &    36.24 &    116.25 \\
    \cline{3-8}
                  &        & \multirow{2}{*}{\textsc{ns}} & \textsc{self-mod} &   120.73 &    125.44 &    33.70 &    101.40 \\
                  &        &       & \textsc{baseline} &   134.13 &    131.89 &    37.12 &    122.74 \\
    \cline{1-8}
    \multirow{8}{*}{\shortstack{\textsc{Spectral}\\\textsc{Norm}}} & \multirow{4}{*}{\textsc{res}} & \multirow{2}{*}{\textsc{hinge}} & \textsc{self-mod} &    14.32 &     24.50 &    18.54 &     68.90 \\
                  &        &       & \textsc{baseline} &    17.10 &     26.15 &    20.08 &     78.62 \\
    \cline{3-8}

                  &        & \multirow{2}{*}{\textsc{ns}} & \textsc{self-mod} &    14.80 &     26.27 &    20.63 &     80.48 \\
                  &        &       & \textsc{baseline} &    17.50 &     30.22 &    23.81 &    120.82 \\
    \cline{2-8}
                  & \multirow{4}{*}{\textsc{sndc}} & \multirow{2}{*}{\textsc{hinge}} & \textsc{self-mod} &    48.07 &     22.51 &    24.66 &     75.87 \\
                  &        &       & \textsc{baseline} &    38.31 &     27.20 &    26.33 &     90.01 \\
    \cline{3-8}
                  &        & \multirow{2}{*}{\textsc{ns}} & \textsc{self-mod} &    46.65 &     24.73 &    26.09 &     76.69 \\
                  &        &       & \textsc{baseline} &    40.80 &     28.16 &    27.41 &     93.25 \\
    \bottomrule
    \multirow{2}{*}{\shortstack{\textsc{Best of above}}} & & & \textsc{self-mod} &   \textbf{14.32} &     \textbf{22.51} &    \textbf{18.54} &     \textbf{68.90} \\
                  &        &       & \textsc{baseline} &    17.10 &     26.15 &    20.08 &     78.62 \\
    \bottomrule
  \end{tabular}
}


\end{table}

\begin{table}[t]
  \centering
  \caption{\label{tab:unconditional_reduction}
    Reduction in FID over a large class of hyperparameter settings,
    losses, regularization, and normalization schemes. We observe from
    4.3\% to 33\% decrease in FID. When applied to the \textsc{resnet}
    architecture, independently of the loss, regularization, and
    normalization, \textsc{self-mod} always outperforms the baseline. For
    \textsc{sndcgan} we observe an improvement in $87.5\%$ of the cases
    (all except two on $\lsun{}{}$).}
  \begin{tabular}{llrr|llrr}
\toprule
         &  & \multicolumn{2}{c}{\textsc{Reduction}(\%)} & & & \multicolumn{2}{c}{\textsc{Reduction}(\%)}\\
         \textsc{Model} &  &    \textsc{resnet} &\textsc{sndc}  & \textsc{Model} & &\textsc{resnet} & \textsc{sndc}  \\

\midrule
\textsc{hinge-gp} & \textsc{bedroom} &     18.50 &    7.30  &  \textsc{ns-gp} & \textsc{bedroom} &     31.22 &    9.99 \\    
         & \textsc{celebahq} &      9.94 &   18.81 &	         & \textsc{celebahq} &     20.96 &    4.89 \\
         & \textsc{cifar10} &      4.30 &    7.33  &	         & \textsc{cifar10} &      6.51 &    9.21 \\ 
         & \textsc{imagenet} &      9.18 &   22.01 &	         & \textsc{imagenet} &     12.92 &   17.39 \\
\textsc{hinge-sn} & \textsc{bedroom} &     16.25 &  -25.48  &	\textsc{ns-sn} & \textsc{bedroom} &     15.43 &  -14.35 \\    
         & \textsc{celebahq} &      6.31 &   17.26 &	         & \textsc{celebahq} &     13.08 &   12.20 \\
         & \textsc{cifar10} &      7.67 &    6.35  &	         & \textsc{cifar10} &     13.36 &    4.83 \\ 
         & \textsc{imagenet} &     12.37 &   15.72 &	         & \textsc{imagenet} &     33.39 &   17.76 \\
  \bottomrule
\end{tabular}

\end{table}

\textbf{Conditional Generation.}
We demonstrate that self-modulation also works for label-conditional
generation.  Here, one is given access the class label which may be
used by the generator and the discriminator.  We compare two settings:
(1) Generator conditioning is applied via label-conditional Batch
Norm~\citep{de2017modulating,miyato2018cgans} with no use of labels in
the discriminator (\textsc{G-Cond}).  (2) Generator conditioning
applied as above, but with projection based conditioning in the
discriminator (intuitively it encourages the discriminator to use label
discriminative features to distinguish true/fake samples), as
in~\cite{miyato2018cgans} (\textsc{P-cGAN}).  The former can be
considered as a special case of the latter where discriminator
conditioning is disabled.  For \textsc{P-cGAN}, we use the
architectures and hyper-parameter settings of~\cite{miyato2018cgans}.
See the appendix, Section~\ref{app:cond} for
details.  In both cases, we compare standard label-conditional
batch normalization to self-modulation with additional labels, as discussed in
Section~\ref{sec:model}, \Eqref{eq:conditioning}.

The results are shown in Table~\ref{tab:conditional_raw}. Again, we
observe that the simple incorporation of self-modulation leads to a
significant improvement in performance in the considered settings.

\begin{table}
  \small
  \centering
  \caption{\label{tab:conditional_raw} FID and IS scores in label conditional setting.}
  \begin{tabular}{lcccccccc}
\toprule
& & \multicolumn{2}{c}{\textsc{Unconditional}} & \multicolumn{2}{c}{\textsc{G-Cond}} & \multicolumn{2}{c}{\textsc{P-cGAN}} \\

&\textsc{Score} & \textsc{Baseline} & \textsc{Self-mod} & \textsc{Baseline} & \textsc{Self-mod} & \textsc{Baseline} & \textsc{Self-mod}\\
\midrule
\textsc{cifar10}  &   \textsc{fid} & 20.41 & \textbf{18.58} & 21.08 & \textbf{18.39} &   16.06 &    \textbf{14.19} \\
\textsc{imagenet} &   \textsc{fid} & 81.07 & \textbf{69.53} & 80.43 & \textbf{68.93} &   70.28 &    \textbf{66.09} \\\midrule
\textsc{cifar10}  &   \textsc{is} &  7.89 &  \textbf{8.31} &  8.11 &  \textbf{8.34} &    8.53 &     \textbf{8.71} \\
\textsc{imagenet} &   \textsc{is} & 11.16 & \textbf{12.52} & 11.16 & \textbf{12.48} &   13.62 &    \textbf{14.14} \\
\bottomrule
\end{tabular}
\end{table}

\textbf{Training for longer on \imagenet.}
To demonstrate that self-modulation continues to yield improvement
after training for longer, we train \imagenet{} for $500$k generator
steps. Due to the increased computational demand we use a single setting for the
unconditional and conditional settings models
following~\citet{miyato2018spectral} and \citet{miyato2018cgans}, but using only two
discriminator steps per generator. We expect that the results would continue to improve if training longer. However, currently results from $500$k
steps require training for $\sim$10 days on a P100 GPU.

We compute the median FID across 3 random seeds. After $500$k steps the
baseline unconditional model attains FID $60.4$, self-modulation attains
$53.7$ ($11\%$ improvement). In the conditional setting self-modulation
improves the FID from $50.6$ to $43.9$ (13\% improvement). The
improvements in IS are from $14.1$ to $15.1$, and $20.1$ to $22.2$ in
unconditional and conditional setting, respectively.

\textbf{Where to apply self-modulation?}
Given the robust improvements
of the proposed method, an immediate question is where to apply the
modulation. We tested two settings: (1) applying modulation to every
batch normalization layer, and (2) applying it to a single layer. The
results of this ablation are in Figure~\ref{fig:fid_main}.
These results suggest
that the benefit of self-modulation is greatest in the last layer, as
may be intuitive, but applying it to each layer is most effective.

\section{Related Work}

\textbf{Conditional GANs.}
Conditioning on side information, such as class labels, has been shown to improve the performance of GANs. Initial proposals were based on concatenating this additional feature with the input vector~\citep{mirza2014conditional,radford2016,odena2017}.
Recent approaches, such as the projection cGAN~\citep{miyato2018cgans} injects label information into the generator architecture using conditional Batch Norm layers~\citep{de2017modulating}. Self-modulation is a simple yet effective complementary addition to this line of work which makes a significant difference when no side information is available. In addition, when side information is available it can be readily applied as discussed in Section~\ref{sec:model} and leads to further improvements.

\textbf{Conditional Modulation.}
Conditional modulation, using side information to modulate the computation flow in neural networks, is a rich idea which has been applied in various contexts (beyond GANs). In particular, \citet{dumoulin2017learned} apply Conditional Instance Normalization~\citep{ulyanov1607instance} to image style-transfer~\citep{dumoulin2017learned}. \citet{kim2017dynamic} use Dynamic Layer Normalization~\citep{ba2016layer} for adaptive acoustic modelling. Feature-wise Linear Modulation~\citep{perez2018} generalizes this family of methods by conditioning the Batch Norm scaling and bias factors (which correspond to multiplicative and additive interactions) on general external embedding vectors in supervised learning. The proposed method applies to generators in GAN (unsupervised learning), and it works with both unconditional (without side information) and conditional (with side information) settings.

\textbf{Multiplicative and Additive Modulation.}
Existing conditional modulations mentioned above are usually instantiated via Batch Normalization, which include both multiplicative and additive modulation. These two types of modulation also link to other techniques widely used in neural network literature. The multiplicative modulation is closely related to Gating, which is adopted in LSTM~\citep{hochreiter1997long}, gated PixelCNN~\citep{van2016conditional}, Convolutional Sequence-to-sequence networks~\citep{gehring2017convolutional} and Squeeze-and-excitation Networks~\citep{hu2018squeeze}. The additive modulation is closely related to Residual Networks~\citep{he2016}. The proposed method adopts both types of modulation.

\section{Discussion}

\begin{figure*}[t]
\centering
\begin{tabular}{m{2cm}m{4cm}m{4cm}}
& $\qquad$ Condition number
& $\qquad$ Precision/Recall \\
\cifar &
\includegraphics[scale=0.32]{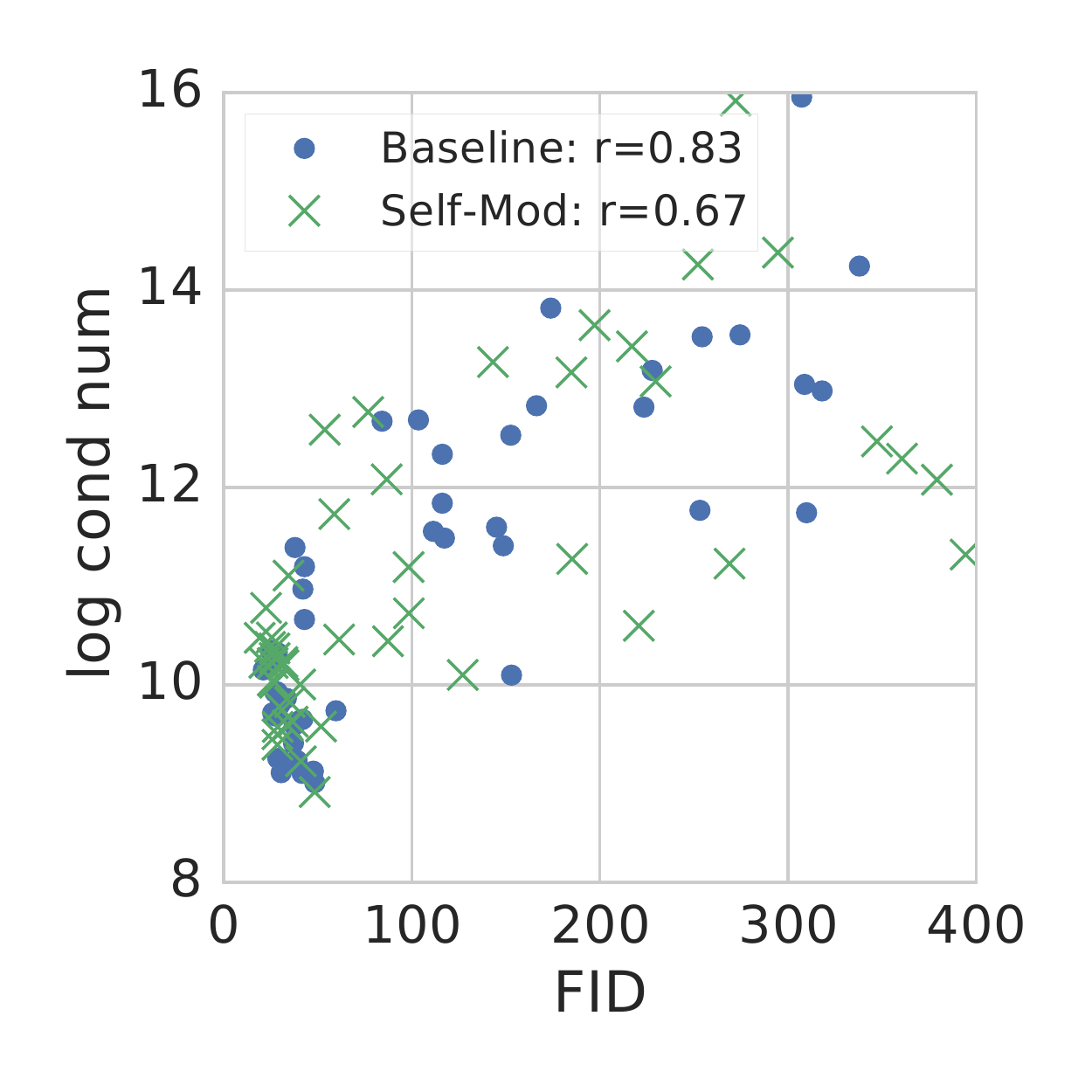} &
\includegraphics[scale=0.32]{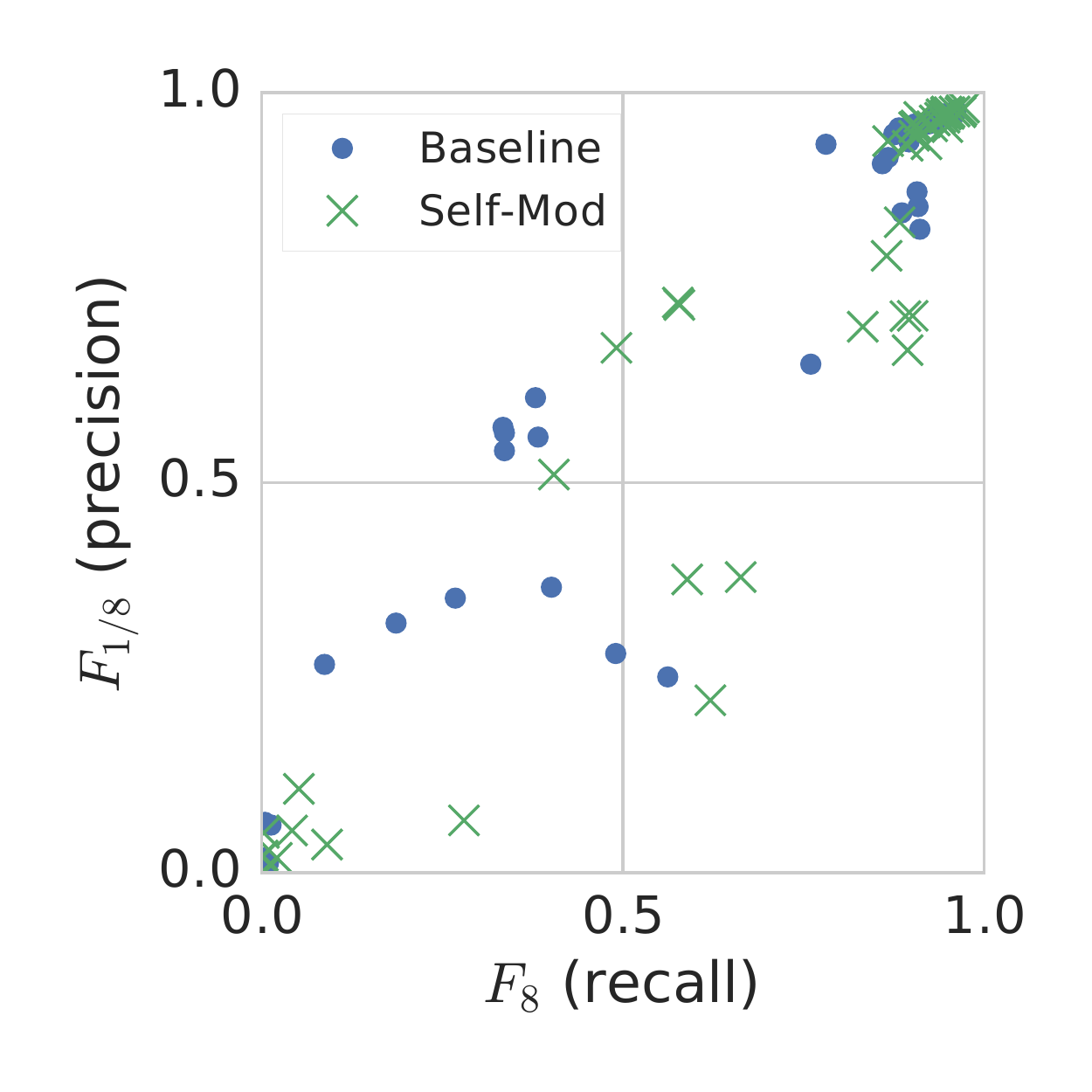} \\
\imagenet &
\includegraphics[scale=0.32]{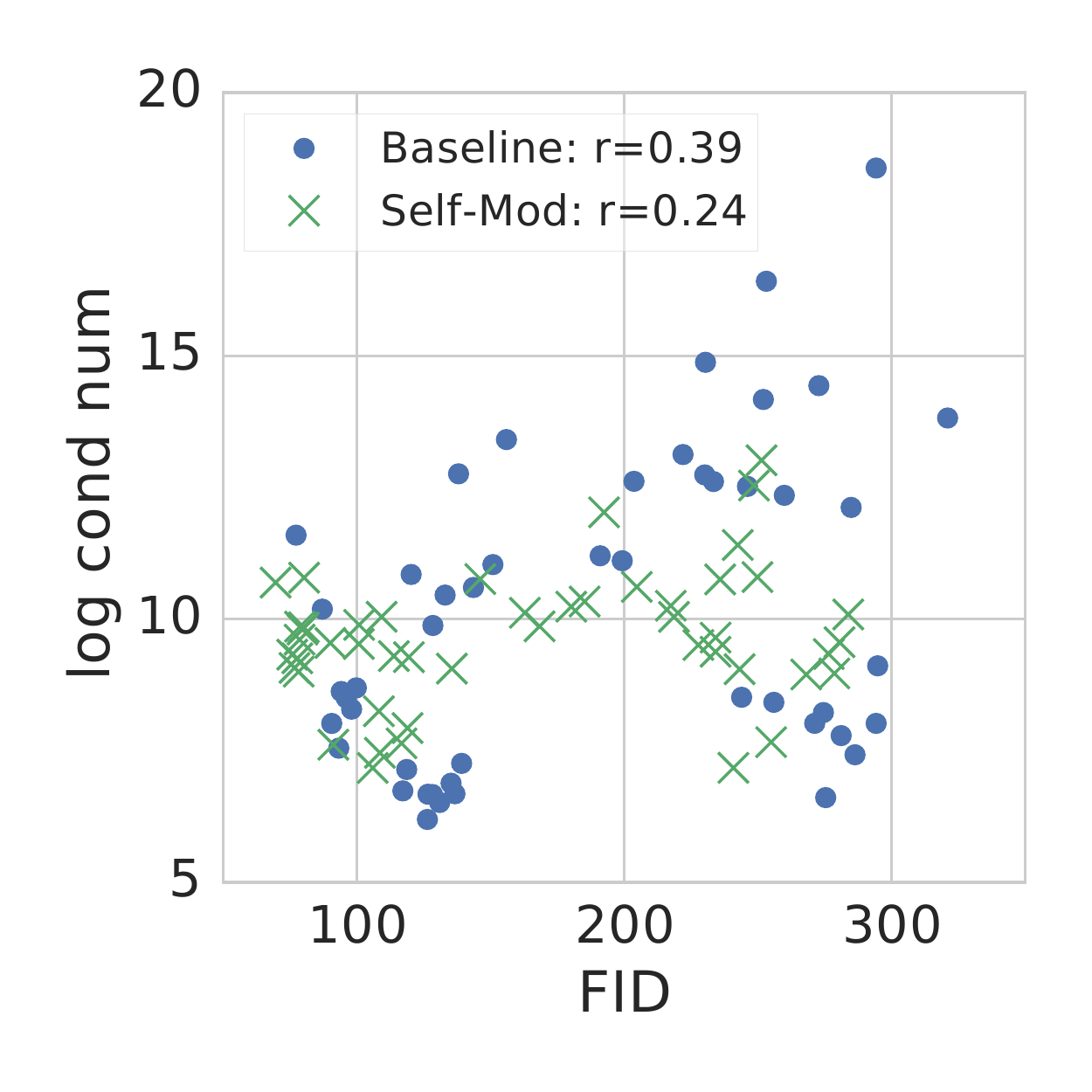} &
\includegraphics[scale=0.32]{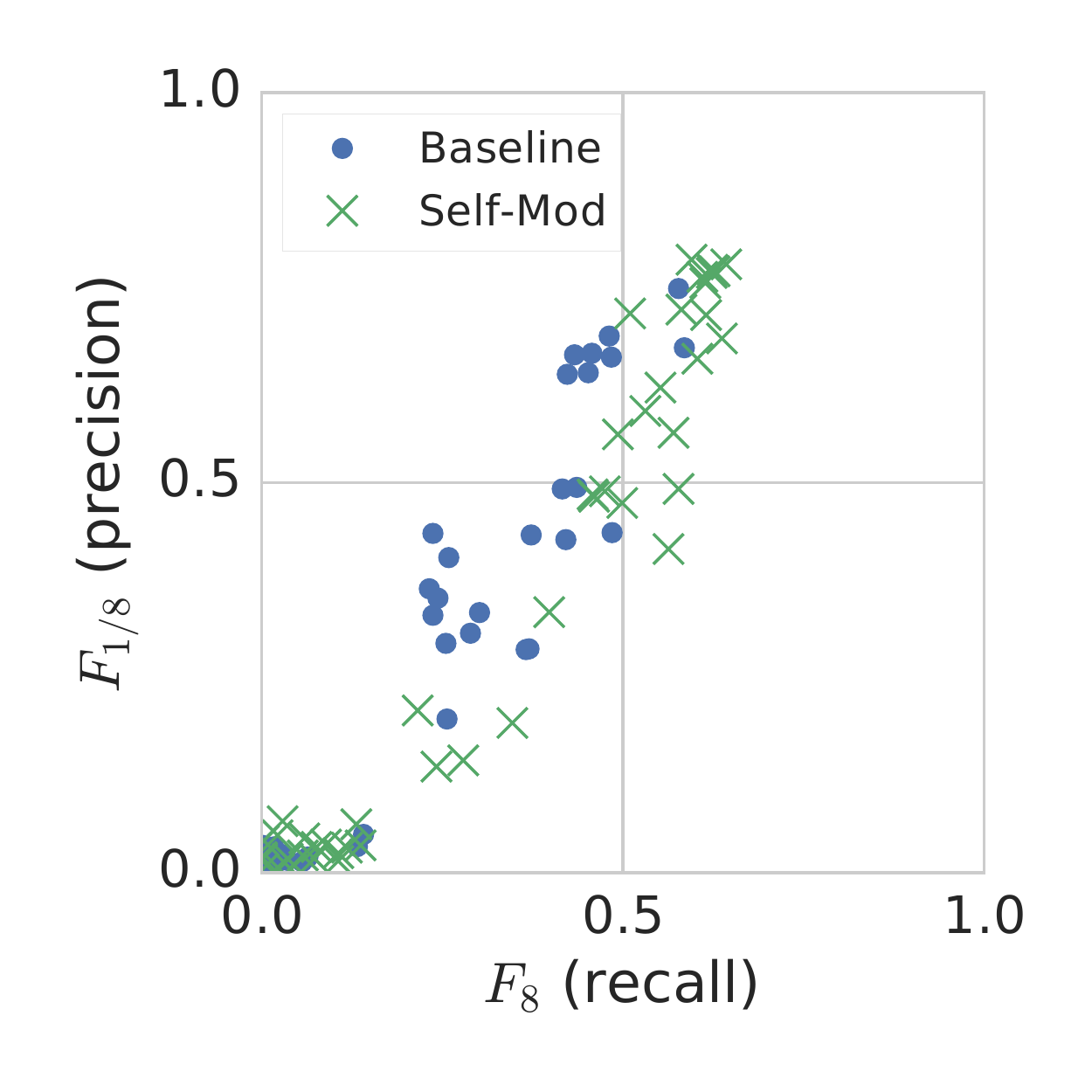}
\end{tabular}
\caption{
Each point corresponds to a single model/hyperparameter setting.
The left-hand plots show the log condition number of the generator versus the FID score.
The right-hand plots show the generator precision/recall curves.
The $r$ values for the correlation between log condition number and FID on \cifar{} are
$0.67$ and $0.83$ for Self-Mod and Base, respectively.
For \imagenet{} they are $0.24$ and $0.39$ for Self-Mod and Base, respectively.
\lsun{} and \celebahq{} are in the appendix.}
\label{fig:condition}
\end{figure*}

We present a generator modification that improves the performance of most GANs.
This technique is simple to implement and can be applied to all popular GANs,
therefore we believe that self-modulation is a useful addition to the GAN toolbox.

Our results suggest that self-modulation clearly yields performance gains,
however, they do not say how this technique results in better models.
Interpretation of deep networks is a complex topic,
especially for GANs, where the training process is less well understood.
Rather than purely speculate, we compute two diagnostic statistics that were proposed recently
ignite the discussion of the method's effects.

First, we compute the condition number of the generators Jacobian.
\citet{odena2018generator} provide evidence that better generators have
a Jacobian with lower condition number and hence regularize using this quantity.
We estimate the generator condition number in the same was as~\citet{odena2018generator}.
We compute the Jacobian $(J_{\bm z})_{i,j} =
\frac{\delta G(\bm z)_i}{\delta z_j}$ at each $\bm z$ in a minibatch,
then average the logarithm of the condition numbers computed from each
Jacobian.

Second, we compute a notion of precision and recall for generative models.
\citet{sajjadi2018assessing} define the quantities, $F_8$ and $F_{1/8}$, for generators.
These quantities relate intuitively to the traditional precision and recall metrics for classification.
Generating points which have low probability under the true data distribution is interpreted as a loss in precision, and is penalized by the $F_8$ score.
Failing to generate points that have high probability under the true data distributions is interpreted as a loss in recall, and is penalized by the $F_{1/8}$ score.

Figure~\ref{fig:condition} shows both statistics.
The left hand plot shows the condition number plotted against FID score for each model.
We observe that poor models tend to have large condition numbers; the correlation,
although noisy, is always positive.
This result corroborates the observations in~\citep{odena2018generator}.
However, we notice an inverse trend in the vicinity of the best models.
The cluster of the best models with self-modulation has lower FID,
but higher condition number, than the best models without self-modulation.
Overall the correlation between FID and condition number is smaller for self-modulated models.
This is surprising, it appears that rather than unilaterally reducing the condition number,
self-modulation provides some training stability, yielding models with a small range of generator condition numbers.

The right-hand plot in Figure~\ref{fig:condition} shows the $F_8$ and $F_{1/8}$ scores.
Models in the upper-left quadrant cover true data modes better (higher precision),
and models in the lower-right quadrant produce more modes (higher recall).
Self-modulated models tend to favor higher recall. This effect is most pronounced on \imagenet{}.

Overall these diagnostics indicate that self-modulation stabilizes the generator
towards favorable conditioning values.
It also appears to improve mode coverage.
However, these metrics are very new; further development of analysis tools and
theoretical study is needed to better disentangle the symptoms and causes
of the self-modulation technique, and indeed of others.

\subsection*{Acknowledgements}

We would like to thank Ilya Tolstikhin for helpful discussions. We would also like to thank Xiaohua
Zhai, Marcin Michalski, Karol Kurach and Anton Raichuk for their help with infustrature.
We also appreciate general discussions with Olivier Bachem, Alexander Kolesnikov,
Thomas Unterthiner, and Josip Djolonga. Finally, we are grateful for the support of other
members of the Google Brain team.

\bibliography{gan_bib}

\begin{thebibliography}{41}
\providecommand{\natexlab}[1]{#1}
\providecommand{\url}[1]{\texttt{#1}}
\expandafter\ifx\csname urlstyle\endcsname\relax
  \providecommand{\doi}[1]{doi: #1}\else
  \providecommand{\doi}{doi: \begingroup \urlstyle{rm}\Url}\fi

\bibitem[Arjovsky et~al.(2017)Arjovsky, Chintala, and
  Bottou]{arjovsky2017wasserstein}
Mart{\'{\i}}n Arjovsky, Soumith Chintala, and L{\'{e}}on Bottou.
\newblock Wasserstein generative adversarial networks.
\newblock In \emph{International Conference on Machine Learning (ICML)}, 2017.

\bibitem[Ba et~al.(2016)Ba, Kiros, and Hinton]{ba2016layer}
Jimmy~Lei Ba, Jamie~Ryan Kiros, and Geoffrey~E Hinton.
\newblock Layer normalization.
\newblock \emph{arXiv preprint arXiv:1607.06450}, 2016.

\bibitem[Barratt \& Sharma(2018)Barratt and Sharma]{barratt2018note}
Shane Barratt and Rishi Sharma.
\newblock A note on the inception score.
\newblock \emph{arXiv preprint arXiv:1801.01973}, 2018.

\bibitem[De~Vries et~al.(2017)De~Vries, Strub, Mary, Larochelle, Pietquin, and
  Courville]{de2017modulating}
Harm De~Vries, Florian Strub, J{\'e}r{\'e}mie Mary, Hugo Larochelle, Olivier
  Pietquin, and Aaron~C Courville.
\newblock Modulating early visual processing by language.
\newblock In \emph{Advances in Neural Information Processing Systems (NIPS)},
  2017.

\bibitem[Dumoulin et~al.(2017)Dumoulin, Shlens, and
  Kudlur]{dumoulin2017learned}
Vincent Dumoulin, Jonathon Shlens, and Manjunath Kudlur.
\newblock A learned representation for artistic style.
\newblock \emph{International Conference on Learning Representations (ICLR)},
  2017.

\bibitem[Gehring et~al.(2017)Gehring, Auli, Grangier, Yarats, and
  Dauphin]{gehring2017convolutional}
Jonas Gehring, Michael Auli, David Grangier, Denis Yarats, and Yann Dauphin.
\newblock Convolutional sequence to sequence learning.
\newblock In \emph{International Conference on Machine Learning (ICML)}, 2017.

\bibitem[Goodfellow et~al.(2014)Goodfellow, Pouget-Abadie, Mirza, Xu,
  Warde-Farley, Ozair, Courville, and Bengio]{goodfellow2014generative}
Ian Goodfellow, Jean Pouget-Abadie, Mehdi Mirza, Bing Xu, David Warde-Farley,
  Sherjil Ozair, Aaron Courville, and Yoshua Bengio.
\newblock Generative adversarial nets.
\newblock In \emph{Advances in Neural Information Processing Systems (NIPS)},
  2014.

\bibitem[Gulrajani et~al.(2017)Gulrajani, Ahmed, Arjovsky, Dumoulin, and
  Courville]{gulrajani2017improved}
Ishaan Gulrajani, Faruk Ahmed, Martin Arjovsky, Vincent Dumoulin, and Aaron
  Courville.
\newblock Improved training of {W}asserstein {GAN}s.
\newblock \emph{Advances in Neural Information Processing Systems (NIPS)},
  2017.

\bibitem[He et~al.(2016)He, Zhang, Ren, and Sun]{he2016}
K.~He, X.~Zhang, S.~Ren, and J.~Sun.
\newblock Deep residual learning for image recognition.
\newblock In \emph{Computer Vision and Pattern Recognition (CVPR)}, 2016.

\bibitem[Heusel et~al.(2017)Heusel, Ramsauer, Unterthiner, Nessler, Klambauer,
  and Hochreiter]{heusel2017gans}
Martin Heusel, Hubert Ramsauer, Thomas Unterthiner, Bernhard Nessler,
  G{\"u}nter Klambauer, and Sepp Hochreiter.
\newblock {GANs trained by a two time-scale update rule converge to a Nash
  equilibrium}.
\newblock In \emph{Advances in Neural Information Processing Systems (NIPS)},
  2017.

\bibitem[Hochreiter \& Schmidhuber(1997)Hochreiter and
  Schmidhuber]{hochreiter1997long}
Sepp Hochreiter and J{\"u}rgen Schmidhuber.
\newblock Long short-term memory.
\newblock \emph{Neural computation}, 1997.

\bibitem[Hu et~al.(2018)Hu, Shen, and Sun]{hu2018squeeze}
Jie Hu, Li~Shen, and Gang Sun.
\newblock Squeeze-and-excitation networks.
\newblock In \emph{Computer Vision and Pattern Recognition (CVPR)}, 2018.

\bibitem[Ioffe \& Szegedy(2015)Ioffe and Szegedy]{ioffe2015batch}
Sergey Ioffe and Christian Szegedy.
\newblock Batch normalization: Accelerating deep network training by reducing
  internal covariate shift.
\newblock \emph{arXiv preprint arXiv:1502.03167}, 2015.

\bibitem[Isola et~al.(2016)Isola, Zhu, Zhou, and Efros]{pix2pix2016}
Phillip Isola, Jun-Yan Zhu, Tinghui Zhou, and Alexei~A Efros.
\newblock Unpaired image-to-image translation using cycle-consistent
  adversarial networks.
\newblock \emph{arxiv}, 2016.

\bibitem[Karras et~al.(2017)Karras, Aila, Laine, and
  Lehtinen]{karras2017progressive}
Tero Karras, Timo Aila, Samuli Laine, and Jaakko Lehtinen.
\newblock Progressive growing of gans for improved quality, stability, and
  variation.
\newblock \emph{Advances in Neural Information Processing Systems (NIPS)},
  2017.

\bibitem[Kim et~al.(2017)Kim, Song, and Bengio]{kim2017dynamic}
Taesup Kim, Inchul Song, and Yoshua Bengio.
\newblock Dynamic layer normalization for adaptive neural acoustic modeling in
  speech recognition.
\newblock In \emph{INTERSPEECH}, 2017.

\bibitem[Kingma \& Ba(2014)Kingma and Ba]{kingma2014adam}
Diederik~P Kingma and Jimmy Ba.
\newblock Adam: A method for stochastic optimization.
\newblock \emph{arXiv preprint arXiv:1412.6980}, 2014.

\bibitem[Kurach et~al.(2018)Kurach, Lucic, Zhai, Michalski, and
  Gelly]{kurach2018}
Karol Kurach, Mario Lucic, Xiaohua Zhai, Marcin Michalski, and Sylvain Gelly.
\newblock The {GAN} {L}andscape: {L}osses, {A}rchitectures, {R}egularization,
  and {N}ormalization.
\newblock \emph{arXiv preprint arXiv:1807.04720}, 2018.

\bibitem[Ledig et~al.(2017)Ledig, Theis, Huszar, Caballero, Cunningham, Acosta,
  Aitken, Tejani, Totz, Wang, et~al.]{ledig2017photo}
Christian Ledig, Lucas Theis, Ferenc Huszar, Jose Caballero, Andrew Cunningham,
  Alejandro Acosta, Andrew Aitken, Alykhan Tejani, Johannes Totz, Zehan Wang,
  et~al.
\newblock Photo-realistic single image super-resolution using a generative
  adversarial network.
\newblock In \emph{Computer Vision and Pattern Recognition (CVPR)}, 2017.

\bibitem[Lucic et~al.(2018)Lucic, Kurach, Michalski, Gelly, and
  Bousquet]{lucic2018}
Mario Lucic, Karol Kurach, Marcin Michalski, Sylvain Gelly, and Olivier
  Bousquet.
\newblock Are {GAN}s {C}reated {E}qual? {A} {L}arge-scale {S}tudy.
\newblock In \emph{Advances in Neural Information Processing Systems (NIPS)},
  2018.

\bibitem[Mao et~al.(2016)Mao, Li, Xie, Lau, Wang, and Smolley]{mao2016least}
Xudong Mao, Qing Li, Haoran Xie, Raymond~YK Lau, Zhen Wang, and Stephen~Paul
  Smolley.
\newblock Least squares generative adversarial networks.
\newblock \emph{International Conference on Computer Vision (ICCV)}, 2016.

\bibitem[Mirza \& Osindero(2014)Mirza and Osindero]{mirza2014conditional}
Mehdi Mirza and Simon Osindero.
\newblock Conditional generative adversarial nets.
\newblock \emph{arXiv preprint arXiv:1411.1784}, 2014.

\bibitem[Miyato \& Koyama(2018)Miyato and Koyama]{miyato2018cgans}
Takeru Miyato and Masanori Koyama.
\newblock cgans with projection discriminator.
\newblock \emph{International Conference on Learning Representations (ICLR)},
  2018.

\bibitem[Miyato et~al.(2018)Miyato, Kataoka, Koyama, and
  Yoshida]{miyato2018spectral}
Takeru Miyato, Toshiki Kataoka, Masanori Koyama, and Yuichi Yoshida.
\newblock Spectral normalization for generative adversarial networks.
\newblock \emph{International Conference on Learning Representations (ICLR)},
  2018.

\bibitem[Odena et~al.(2017)Odena, Olah, and Shlens]{odena2017}
Augustus Odena, Christopher Olah, and Jonathon Shlens.
\newblock Conditional image synthesis with auxiliary classifier {GAN}s.
\newblock In \emph{International Conference on Machine Learning (ICML)}, 2017.

\bibitem[Odena et~al.(2018)Odena, Buckman, Olsson, Brown, Olah, Raffel, and
  Goodfellow]{odena2018generator}
Augustus Odena, Jacob Buckman, Catherine Olsson, Tom~B Brown, Christopher Olah,
  Colin Raffel, and Ian Goodfellow.
\newblock Is generator conditioning causally related to gan performance?
\newblock \emph{arXiv preprint arXiv:1802.08768}, 2018.

\bibitem[Pathak et~al.(2016)Pathak, Krahenbuhl, Donahue, Darrell, and
  Efros]{pathak2016context}
Deepak Pathak, Philipp Krahenbuhl, Jeff Donahue, Trevor Darrell, and Alexei~A
  Efros.
\newblock Context encoders: Feature learning by inpainting.
\newblock In \emph{Computer Vision and Pattern Recognition (CVPR)}, 2016.

\bibitem[Perez et~al.(2018)Perez, Strub, de~Vries, Dumoulin, and
  Courville]{perez2018}
Ethan Perez, Florian Strub, Harm de~Vries, Vincent Dumoulin, and Aaron~C.
  Courville.
\newblock Film: Visual reasoning with a general conditioning layer.
\newblock \emph{AAAI}, 2018.

\bibitem[Radford et~al.(2016)Radford, Metz, and Chintala]{radford2016}
Alec Radford, Luke Metz, and Soumith Chintala.
\newblock Unsupervised representation learning with deep convolutional
  generative adversarial networks.
\newblock \emph{International Conference on Learning Representations (ICLR)},
  2016.

\bibitem[Sajjadi et~al.(2018)Sajjadi, Bachem, Lucic, Bousquet, and
  Gelly]{sajjadi2018assessing}
Mehdi~SM Sajjadi, Olivier Bachem, Mario Lucic, Olivier Bousquet, and Sylvain
  Gelly.
\newblock Assessing generative models via precision and recall.
\newblock In \emph{Advances in Neural Information Processing Systems (NIPS)},
  2018.

\bibitem[Salimans et~al.(2016)Salimans, Goodfellow, Zaremba, Cheung, Radford,
  and Chen]{salimans2016improved}
Tim Salimans, Ian Goodfellow, Wojciech Zaremba, Vicki Cheung, Alec Radford, and
  Xi~Chen.
\newblock Improved techniques for training gans.
\newblock In \emph{Advances in Neural Information Processing Systems (NIPS)},
  2016.

\bibitem[Szegedy et~al.(2015)Szegedy, Liu, Jia, Sermanet, Reed, Anguelov,
  Erhan, Vanhoucke, and Rabinovich]{szegedy2015going}
Christian Szegedy, Wei Liu, Yangqing Jia, Pierre Sermanet, Scott Reed, Dragomir
  Anguelov, Dumitru Erhan, Vincent Vanhoucke, and Andrew Rabinovich.
\newblock Going deeper with convolutions.
\newblock In \emph{Computer Vision and Pattern Recognition (CVPR)}, 2015.

\bibitem[Tschannen et~al.(2018)Tschannen, Agustsson, and
  Lucic]{tschannen2018distributionpreserving}
Michael Tschannen, Eirikur Agustsson, and Mario Lucic.
\newblock Deep generative models for distribution-preserving lossy compression.
\newblock In \emph{Advances in Neural Information Processing Systems (NIPS)},
  2018.

\bibitem[Ulyanov et~al.(2016)Ulyanov, Vedaldi, and
  Lempitsky]{ulyanov1607instance}
D~Ulyanov, A~Vedaldi, and VS~Lempitsky.
\newblock Instance normalization: The missing ingredient for fast stylization.
\newblock \emph{arXiv preprint arXiv:1607.08022}, 2016.

\bibitem[van~den Oord et~al.(2016)van~den Oord, Kalchbrenner, Espeholt,
  Vinyals, Graves, et~al.]{van2016conditional}
Aaron van~den Oord, Nal Kalchbrenner, Lasse Espeholt, Oriol Vinyals, Alex
  Graves, et~al.
\newblock Conditional image generation with pixelcnn decoders.
\newblock In \emph{Advances in Neural Information Processing Systems (NIPS)},
  2016.

\bibitem[Vaswani et~al.(2017)Vaswani, Shazeer, Parmar, Uszkoreit, Jones, Gomez,
  Kaiser, and Polosukhin]{vaswani2017attention}
Ashish Vaswani, Noam Shazeer, Niki Parmar, Jakob Uszkoreit, Llion Jones,
  Aidan~N Gomez, {\L}ukasz Kaiser, and Illia Polosukhin.
\newblock Attention is all you need.
\newblock In \emph{Advances in Neural Information Processing Systems (NIPS)},
  2017.

\bibitem[Yu et~al.(2015)Yu, Zhang, Song, Seff, and Xiao]{yu15lsun}
Fisher Yu, Yinda Zhang, Shuran Song, Ari Seff, and Jianxiong Xiao.
\newblock Lsun: Construction of a large-scale image dataset using deep learning
  with humans in the loop.
\newblock \emph{arXiv preprint arXiv:1506.03365}, 2015.

\bibitem[Zhang et~al.(2017)Zhang, Xu, Li, Zhang, Huang, Wang, and
  Metaxas]{zhang2017stackgan}
Han Zhang, Tao Xu, Hongsheng Li, Shaoting Zhang, Xiaolei Huang, Xiaogang Wang,
  and Dimitris Metaxas.
\newblock Stackgan: Text to photo-realistic image synthesis with stacked
  generative adversarial networks.
\newblock \emph{International Conference on Computer Vision (ICCV)}, 2017.

\bibitem[Zhang et~al.(2018)Zhang, Goodfellow, Metaxas, and
  Odena]{zhang2018self}
Han Zhang, Ian Goodfellow, Dimitris Metaxas, and Augustus Odena.
\newblock Self-attention generative adversarial networks.
\newblock \emph{arXiv preprint arXiv:1805.08318}, 2018.

\bibitem[Zhou et~al.(2018)Zhou, Song, Yu, and Yu]{zhou2018understanding}
Zhiming Zhou, Yuxuan Song, Lantao Yu, and Yong Yu.
\newblock Understanding the effectiveness of lipschitz constraint in training
  of gans via gradient analysis.
\newblock \emph{arXiv preprint arXiv:1807.00751}, 2018.

\bibitem[Zhu et~al.(2017)Zhu, Park, Isola, and Efros]{zhu2017unpaired}
Jun-Yan Zhu, Taesung Park, Phillip Isola, and Alexei~A Efros.
\newblock Unpaired image-to-image translation using cycle-consistent
  adversarial networks.
\newblock \emph{arXiv preprint}, 2017.

\end{thebibliography}
\bibliographystyle{iclr2019_conference}

\newpage

\appendix

\section{Additional results}

\subsection{Inception Scores}

\begin{table}[h!]
  \centering
  \caption{\label{tab:unconditional_is_raw}
  In the unpaired setting (as defined in Section~\ref{sec:experiments_unconditional}),
  we compute the median score (across random seeds) and report the best
  attainable score across considered optimization hyperparameters.
  \textsc{Self-Mod} is the method introduced in Section~\ref{sec:model} and
  \textsc{baseline} refers to batch normalization. }
  \scriptsize
  {\renewcommand{\arraystretch}{1.2}
  \begin{tabular}{llllrrrr}
    \toprule
    \textsc{Type} & \textsc{Arch} & \textsc{Loss} & \textsc{Method} & \textsc{bedroom} & \textsc{celebahq} & \textsc{cifar10} & \textsc{imagenet} \\
    \midrule
    \multirow{8}{*}{\shortstack{\textsc{Gradient}\\\textsc{penalty}}} & \multirow{4}{*}{\textsc{resnet}} & \multirow{2}{*}{\textsc{hinge}} & \textsc{self-mod} & 5.28 $\pm$ 0.18 & 2.92 $\pm$ 0.13 & 7.71 $\pm$ 0.59 & 11.52 $\pm$ 0.07 \\
                  & & & \textsc{baseline} & 4.72 $\pm$ 0.11 & 2.80 $\pm$ 0.08 & 7.35 $\pm$ 0.02 & 10.26 $\pm$ 0.09 \\
    \cline{3-8} 
                  & & \multirow{2}{*}{\textsc{ns}} & \textsc{self-mod} & 4.96 $\pm$ 0.17 & 2.61 $\pm$ 0.05 & 7.70 $\pm$ 0.05 & 10.74 $\pm$ 1.20 \\
                  & & & \textsc{baseline} & 4.54 $\pm$ 0.11 & 2.60 $\pm$ 0.25 & 7.26 $\pm$ 0.03 & 9.49 $\pm$ 0.12 \\
    \cline{2-8} 
    \cline{3-8} 
                  & \multirow{4}{*}{\textsc{sndcgan}} & \multirow{2}{*}{\textsc{hinge}} & \textsc{self-mod} & 6.34 $\pm$ 0.07 & 3.05 $\pm$ 0.12 & 7.37 $\pm$ 0.04 & 10.99 $\pm$ 0.06 \\
                  & & & \textsc{baseline} & 5.02 $\pm$ 0.05 & 3.08 $\pm$ 0.09 & 6.88 $\pm$ 0.05 & 8.11 $\pm$ 0.06 \\
    \cline{3-8} 
                  & & \multirow{2}{*}{\textsc{ns}} & \textsc{self-mod} & 6.31 $\pm$ 0.05 & 3.07 $\pm$ 0.05 & 7.28 $\pm$ 0.06 & 10.06 $\pm$ 0.10 \\
                  & & & \textsc{baseline} & 4.71 $\pm$ 0.05 & 3.21 $\pm$ 0.20 & 6.86 $\pm$ 0.06 & 7.24 $\pm$ 0.16 \\
    \cline{1-8} 
    \cline{2-8} 
    \cline{3-8} 
    \multirow{8}{*}{\shortstack{\textsc{Spectral}\\\textsc{Norm}}} & \multirow{4}{*}{\textsc{resnet}} & \multirow{2}{*}{\textsc{hinge}} & \textsc{self-mod} & 3.94 $\pm$ 0.22 & 3.65 $\pm$ 0.16 & 8.29 $\pm$ 0.03 & 12.67 $\pm$ 0.07 \\
                  & & & \textsc{baseline} & 4.32 $\pm$ 0.17 & 3.26 $\pm$ 0.16 & 8.00 $\pm$ 0.03 & 11.29 $\pm$ 0.12 \\
    \cline{3-8} 
                  & & \multirow{2}{*}{\textsc{ns}} & \textsc{self-mod} & 4.61 $\pm$ 0.18 & 3.32 $\pm$ 0.09 & 8.23 $\pm$ 0.04 & 11.52 $\pm$ 0.28 \\
                  & & & \textsc{baseline} & 4.07 $\pm$ 0.21 & 2.58 $\pm$ 0.08 & 7.93 $\pm$ 0.04 & 7.40 $\pm$ 0.60 \\
    \cline{2-8} 
    \cline{3-8} 
                  & \multirow{4}{*}{\textsc{sndcgan}} & \multirow{2}{*}{\textsc{hinge}} & \textsc{self-mod} & 5.85 $\pm$ 0.07 & 2.74 $\pm$ 0.02 & 7.90 $\pm$ 0.04 & 12.50 $\pm$ 0.12 \\
                  & & & \textsc{baseline} & 4.82 $\pm$ 0.12 & 2.40 $\pm$ 0.02 & 7.48 $\pm$ 0.04 & 9.62 $\pm$ 0.10 \\
    \cline{3-8} 
                  & & \multirow{2}{*}{\textsc{ns}} & \textsc{self-mod} & 5.73 $\pm$ 0.07 & 2.55 $\pm$ 0.02 & 7.84 $\pm$ 0.02 & 11.95 $\pm$ 0.09 \\
                  & & & \textsc{baseline} & 4.39 $\pm$ 0.14 & 2.33 $\pm$ 0.01 & 7.37 $\pm$ 0.04 & 9.28 $\pm$ 0.13 \\
    \bottomrule 
  \end{tabular}
}

\end{table}

\subsection{FIDs}
\begin{table}[h]
  \centering
  \caption{\label{tab:unconditional_std}
    Table~\ref{tab:unconditional_raw} with the standard error of the median.}
  \small
  {\renewcommand{\arraystretch}{1.2}
  \scriptsize
  \begin{tabular}{llllrrrr}
    \toprule
    \textsc{Type} & \textsc{Arch} & \textsc{Loss} & \textsc{Method} & \textsc{bedroom} & \textsc{celebahq} & \textsc{cifar10} & \textsc{imagenet}\\
    \midrule
    \multirow{8}{*}{\shortstack{\textsc{Gradient}\\\textsc{penalty}}} & \multirow{4}{*}{\textsc{res}}& \multirow{2}{*}{\textsc{hinge}} & \textsc{self-mod} & 22.62 $\pm$ 64.79 & 27.03 $\pm$ 0.29 & 26.93 $\pm$ 13.52 & 78.31 $\pm$ 0.96 \\
                  && & \textsc{base} & 27.75 $\pm$ 1.01 & 30.02 $\pm$ 0.69 & 28.14 $\pm$ 0.52 & 86.23 $\pm$ 1.34 \\
    \cline{3-8} 
                  && \multirow{2}{*}{\textsc{ns}} & \textsc{self-mod} & 25.30 $\pm$ 1.21 & 26.65 $\pm$ 13.16 & 26.74 $\pm$ 0.42 & 85.67 $\pm$ 11.94 \\
                  && & \textsc{base} & 36.79 $\pm$ 0.25 & 33.72 $\pm$ 0.78 & 28.61 $\pm$ 0.27 & 98.38 $\pm$ 1.48 \\
    \cline{2-8} 
    \cline{3-8} 
                  & \multirow{4}{*}{\textsc{sndc}} & \multirow{2}{*}{\textsc{hinge}} & \textsc{self-mod} & 110.86 $\pm$ 1.72 & 55.63 $\pm$ 0.53 & 33.58 $\pm$ 0.47 & 90.67 $\pm$ 0.49 \\
                  && & \textsc{base} & 119.59 $\pm$ 1.71 & 68.51 $\pm$ 1.66 & 36.24 $\pm$ 0.69 & 116.25 $\pm$ 0.48 \\
    \cline{3-8} 
                  && \multirow{2}{*}{\textsc{ns}} & \textsc{self-mod} & 120.73 $\pm$ 2.10 & 125.44 $\pm$ 11.27 & 33.70 $\pm$ 0.47 & 101.40 $\pm$ 1.17 \\
                  && & \textsc{base} & 134.13 $\pm$ 2.40 & 131.89 $\pm$ 42.16 & 37.12 $\pm$ 0.62 & 122.74 $\pm$ 0.58 \\
    \cline{1-8} 
    \cline{2-8} 
    \cline{3-8} 
    \multirow{8}{*}{\shortstack{\textsc{Spectral}\\\textsc{Norm}}} & \multirow{4}{*}{\textsc{res}} & \multirow{2}{*}{\textsc{hinge}} & \textsc{self-mod} & 14.32 $\pm$ 0.40 & 24.50 $\pm$ 0.46 & 18.54 $\pm$ 0.15 & 68.90 $\pm$ 0.67 \\
                  && & \textsc{base} & 17.10 $\pm$ 1.44 & 26.15 $\pm$ 0.70 & 20.08 $\pm$ 0.31 & 78.62 $\pm$ 0.97 \\
    \cline{3-8} 
                  && \multirow{2}{*}{\textsc{ns}} & \textsc{self-mod} & 14.80 $\pm$ 0.40 & 26.27 $\pm$ 0.48 & 20.63 $\pm$ 0.20 & 80.48 $\pm$ 2.43 \\
                  && & \textsc{base} & 17.50 $\pm$ 0.64 & 30.22 $\pm$ 0.48 & 23.81 $\pm$ 0.17 & 120.82 $\pm$ 6.82 \\
    \cline{2-8} 
    \cline{3-8} 
                  & \multirow{4}{*}{\textsc{sndc}} & \multirow{2}{*}{\textsc{hinge}} & \textsc{self-mod} & 48.07 $\pm$ 1.77 & 22.51 $\pm$ 0.38 & 24.66 $\pm$ 0.40 & 75.87 $\pm$ 0.37 \\
                  && & \textsc{base} & 38.31 $\pm$ 1.42 & 27.20 $\pm$ 0.80 & 26.33 $\pm$ 0.54 & 90.01 $\pm$ 1.06 \\
    \cline{3-8} 
                  && \multirow{2}{*}{\textsc{ns}} & \textsc{self-mod} & 46.65 $\pm$ 2.72 & 24.73 $\pm$ 0.25 & 26.09 $\pm$ 0.19 & 76.69 $\pm$ 0.89 \\
                  && & \textsc{base} & 40.80 $\pm$ 1.75 & 28.16 $\pm$ 0.17 & 27.41 $\pm$ 0.43 & 93.25 $\pm$ 0.35 \\
    \bottomrule
    \multirow{2}{*}{\shortstack{\textsc{Best of above}}} && & \textsc{self-mod} & \textbf{14.32} & \textbf{22.51} & \textbf{18.54} & \textbf{68.90} \\
                  && & \textsc{baseline} & 17.10 & 26.15 & 20.08 & 78.62 \\
    \bottomrule
  \end{tabular}
}


\end{table}

\newpage

\subsection{Which layer to modulate?}
\begin{figure}[ht]
\begin{center}
\includegraphics[width=\textwidth]{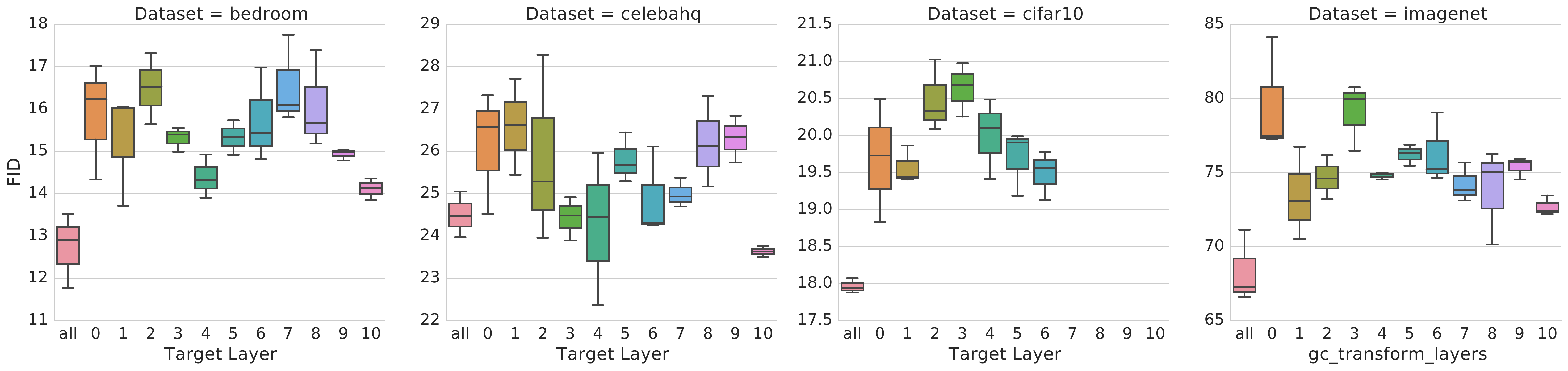}
\end{center}
\caption{\label{fig:which_layer}
FID distributions resulting from Self-Modulation on different layers.}
\end{figure}

Figure~\ref{fig:which_layer} presents the performance when modulating different layers of the generator for each dataset.

\subsection{Conditioning and Precision/Recall}

\begin{figure*}[ht]
\centering
\begin{tabular}{m{2cm}m{6cm}m{6cm}}
\cifar: &
\includegraphics[scale=0.35]{figure/f_c_cifar10.pdf} &
\includegraphics[scale=0.35]{figure/pr_cifar10.pdf} \\
\imagenet: &
\includegraphics[scale=0.35]{figure/f_c_imagenet.pdf} &
\includegraphics[scale=0.35]{figure/pr_imagenet.pdf} \\
\lsun: &
\includegraphics[scale=0.35]{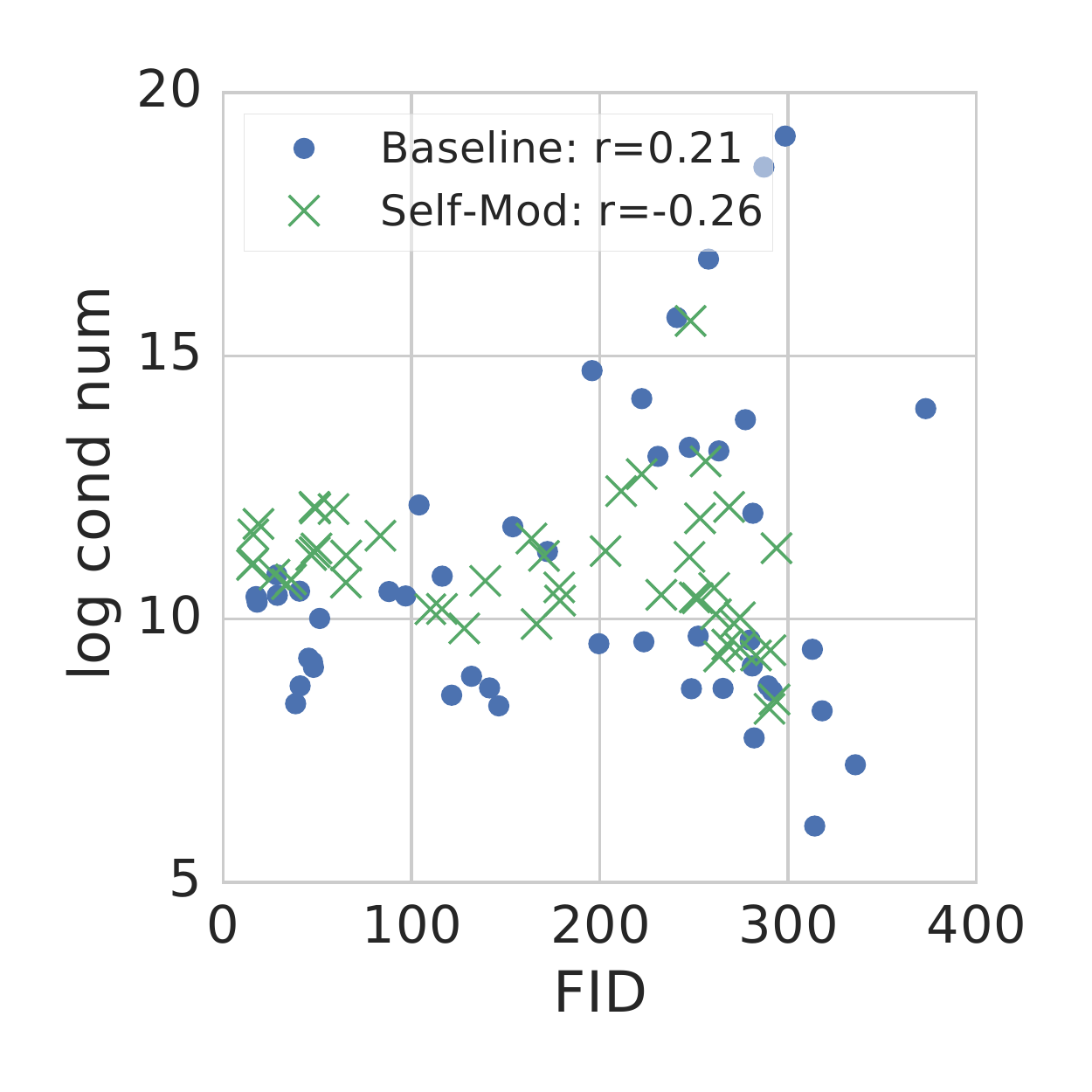} &
\includegraphics[scale=0.35]{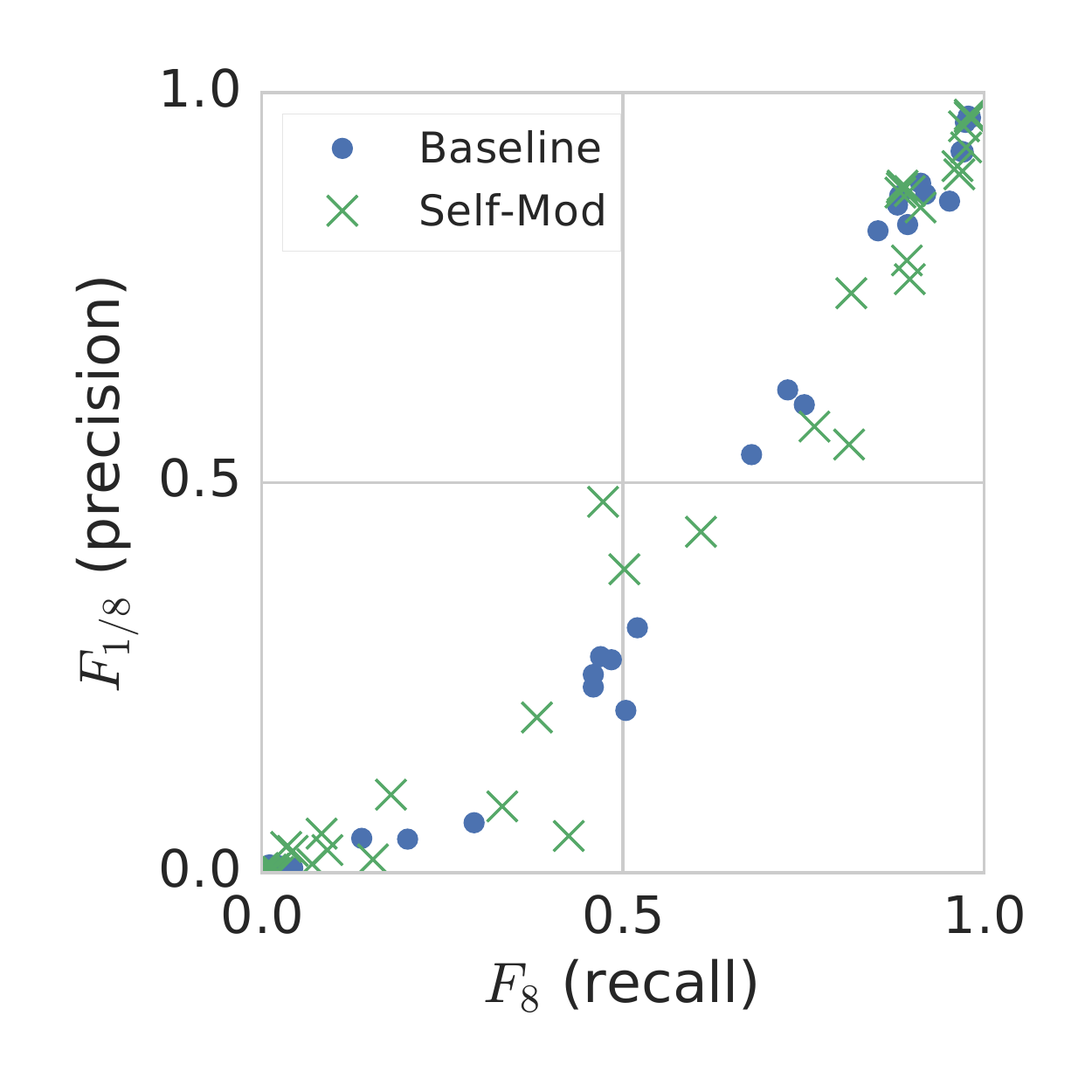} \\
\celebahq: &
\includegraphics[scale=0.35]{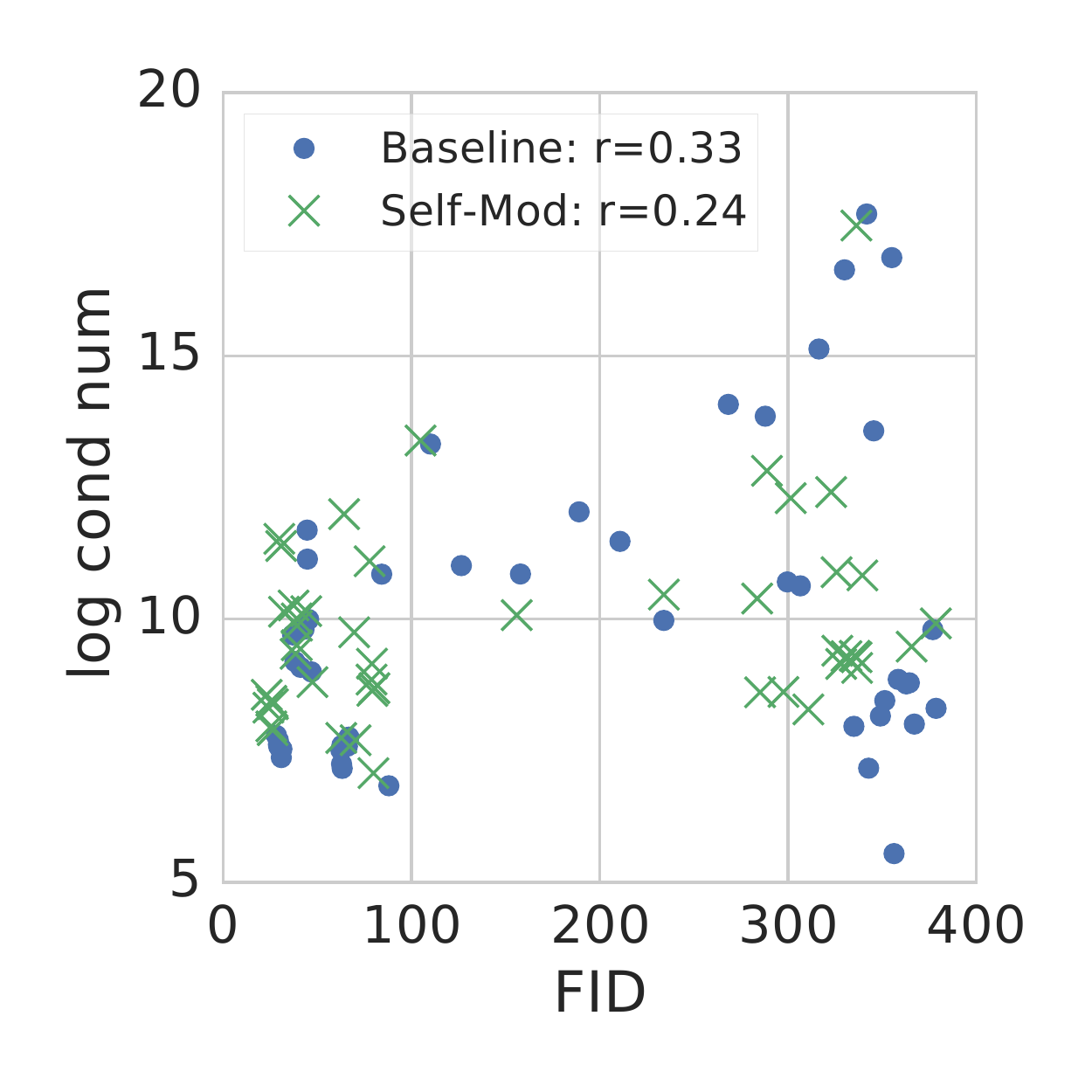} &
\includegraphics[scale=0.35]{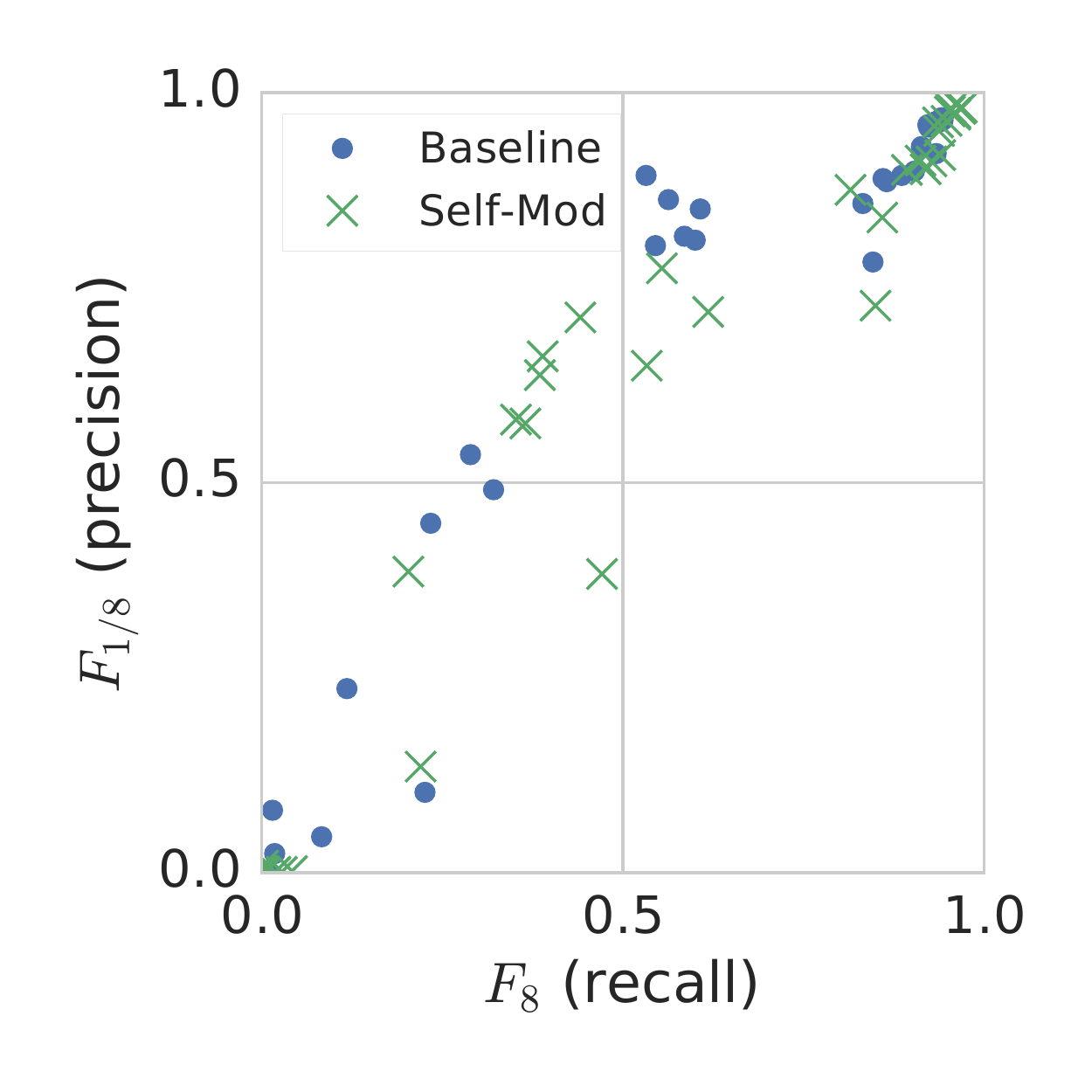}
\end{tabular}
\caption{
Each point in each plot corresponds to a single model for all parameter configurations.
The model with mean FID score across the five random seeds was chosen.
The left-hand plots show the log condition number of the generator versus the FID score for each model.
The right-hand generator precision/recall metrics.
\label{fig:condition_full}
}
\end{figure*}

Figure~\ref{fig:condition_full} presents the generator Jacobian condition number and precision/recall plot for each dataset.

\section{Model Architectures}
\label{sec:app_model_arch}

We describe the model structures that are used in our experiments in this section.

\subsection{SNDCGAN Architectures}

The SNDCGAN architecture we follows the ones used in~\cite{miyato2018spectral}. Since the resolution of images in \cifar is $32\times32\times3$, while resolutions of images in other datasets are $128\times128\times3$. There are slightly differences in terms of spatial dimensions for both architectures. The proposed self-modulation is applied to replace existing BN layer, we term it sBN (self-modulated BN) for short in Table \ref{tab:sndcgan_g_small}, \ref{tab:sndcgan_d_small}, \ref{tab:sndcgan_g_large}, \ref{tab:sndcgan_d_large}.

\subsection{ResNet Architectures}

The ResNet architecture we also follows the ones used in~\cite{miyato2018spectral}. Again, due to the resolution differences, two ResNet architectures are used in this work. The proposed self-modulation is applied to replace existing BN layer, we term it sBN (self-modulated BN) for short in Table \ref{tab:resnet_g_small}, \ref{tab:resnet_d_small}, \ref{tab:resnet_g_large}, \ref{tab:resnet_d_large}.

\subsection{Conditional GAN Architecture}\label{app:cond}

For the conditional setting with label information available, we adopt the Projection Based Conditional GAN (P-cGAN)~\citep{miyato2018cgans}.  There are both conditioning in generators as well ad discriminators. For generator, conditional batch norm is applied via conditioning on label information, more specifically, this can be expressed as follows,
$$
\bm h'_\ell = \bm\gamma_y \odot \frac{\bm h_\ell - \bm \mu}{\bm \sigma} + \bm \beta_y
$$
Where each label $y$ is associated with a scaling and shifting parameters independently.

For discriminator label conditioning, the dot product between final layer feature $\bm\phi(\bm x)$ and label embedding $\mathrm E(y)$ is added back to the discriminator output logits, i.e. $D(\bm x, y) = \bm \psi (\bm\phi(\bm x))+ \bm \phi(\bm x)^T \mathrm E(y)$ where $\bm \phi(\bm x)$ represents the final feature representation layer of input $\bm x$, and $\bm \psi(\cdot)$ is the linear transformation maps the feature vector into a real number. Intuitively, this type of conditional discriminator encourages discriminator to use label discriminative features to distinguish true/fake samples. Both the above conditioning strategies do not dependent on the specific architectures, and can be applied to above architectures with small modifications.

We use the same architectures and hyper-parameter settings\footnote{With one exception: to make it consistent with previous unconditional settings (and also due to the computation time), instead of running five discriminator steps per generator step, we only use two discriminator steps per generator step.} as in \cite{miyato2018cgans}. More specifically, the architecture is the same as ResNet above, and we compare in two settings: (1) only generator label conditioning is applied, and there is no projection based conditioning in the discriminator, and (2) both generator and discriminator conditioning are applied, which is the standard full P-cGAN.


\begin{table}[h]
\centering
\small
\caption{\label{tab:sndcgan_g_small}SNDCGAN Generator with $32\times32\times3$ resolution. sBN denotes BN with self-modulation as proposed.}
\begin{tabular}{lcc}
\hline\hline
Layer           & Details & Output size \\ \hline
Latent noise &  $\bm z\sim \mathcal{N}(0, I)$& $128$            \\\hline
Fully Connected &      Linear        &      $2\cdot2\cdot512$    \\
& Reshape         &     $2\times2\times512$     \\ \hline
Deconv&     sBN, ReLU         &    $2\times2\times512$      \\
&     Deconv4x4,stride=2         &    $4\times4\times256$      \\ \hline
Deconv&     sBN, ReLU         &    $4\times4\times256$      \\
&     Deconv4x4,stride=2         &    $8\times8\times128$      \\ \hline
Deconv&     sBN, ReLU         &    $8\times8\times128$      \\
&     Deconv4x4,stride=2         &    $16\times16\times64$      \\ \hline
Deconv&     sBN, ReLU         &    $16\times16\times64$      \\
&     Deconv4x4,stride=2         &    $32\times32\times3$      \\
&     Tanh        &    $32\times32\times3$      \\ \hline
\hline
\end{tabular}
\end{table}

\begin{table}[]
\centering
\small
\caption{\label{tab:sndcgan_d_small}SNDCGAN Discriminator with $32\times32\times3$ resolution.}
\begin{tabular}{lcc}
\hline\hline
Layer           & Details & Output size \\ \hline
Input image&      -        &      $32\times32\times3$    \\ \hline
Conv&     Conv3x3,stride=1        &    $32\times32\times64$      \\ 
&     LeakyReLU        &    $32\times32\times64$      \\ \hline
Conv&     Conv4x4,stride=2        &    $16\times16\times128$      \\ 
&     LeakyReLU        &    $16\times16\times128$      \\ \hline
Conv&     Conv3x3,stride=1        &    $16\times16\times128$      \\ 
&     LeakyReLU        &    $16\times16\times128$      \\ \hline
Conv&     Conv4x4,stride=2        &    $8\times8\times256$      \\ 
&     LeakyReLU        &    $8\times8\times256$      \\ \hline
Conv&     Conv3x3,stride=1        &    $8\times8\times256$      \\ 
&     LeakyReLU        &    $8\times8\times256$      \\ \hline
Conv&     Conv4x4,stride=2        &    $4\times4\times512$      \\ 
&     LeakyReLU        &    $4\times4\times512$      \\ \hline
Conv&     Conv3x3,stride=1        &    $4\times4\times512$      \\ 
&     LeakyReLU        &    $4\times4\times512$      \\ \hline
Fully connected&     Reshape       &    $4\cdot4\cdot512$      \\ 
&     Linear        &    $1$      \\ \hline
\hline
\end{tabular}
\end{table}

\begin{table}[]
\centering
\small
\caption{\label{tab:sndcgan_g_large}SNDCGAN Gnerator with $128\times128\times3$ resolution. sBN denotes BN with self-modulation as proposed.}
\begin{tabular}{lcc}
\hline\hline
Layer           & Details & Output size \\ \hline
Latent noise &  $\bm z\sim \mathcal{N}(0, I)$& $128$            \\\hline
Fully Connected &      Linear        &      $8\cdot8\cdot512$    \\
& Reshape         &     $8\times8\times512$     \\ \hline
Deconv&     sBN, ReLU         &    $8\times8\times512$      \\
&     Deconv4x4,stride=2         &    $16\times16\times256$      \\ \hline
Deconv&     sBN, ReLU         &    $16\times16\times256$      \\
&     Deconv4x4,stride=2         &    $32\times32\times128$      \\ \hline
Deconv&     sBN, ReLU         &    $32\times32\times128$      \\
&     Deconv4x4,stride=2         &    $64\times64\times64$      \\ \hline
Deconv&     sBN, ReLU         &    $64\times64\times64$      \\
&     Deconv4x4,stride=2         &    $128\times128\times3$      \\
&     Tanh        &    $128\times128\times3$      \\ \hline
\hline
\end{tabular}
\end{table}

\begin{table}[]
\centering
\small
\caption{\label{tab:sndcgan_d_large}SNDCGAN Discriminator with $128\times128\times3$ resolution.}
\begin{tabular}{lcc}
\hline\hline
Layer           & Details & Output size \\ \hline
Input image&      -        &      $128\times128\times3$    \\ \hline
Conv&     Conv3x3,stride=1        &    $128\times128\times64$      \\ 
&     LeakyReLU        &    $128\times128\times64$      \\ \hline
Conv&     Conv4x4,stride=2        &    $64\times64\times128$      \\ 
&     LeakyReLU        &    $64\times64\times128$      \\ \hline
Conv&     Conv3x3,stride=1        &    $64\times64\times128$      \\ 
&     LeakyReLU        &    $64\times64\times128$      \\ \hline
Conv&     Conv4x4,stride=2        &    $32\times32\times256$      \\ 
&     LeakyReLU        &    $32\times32\times256$      \\ \hline
Conv&     Conv3x3,stride=1        &    $32\times32\times256$      \\ 
&     LeakyReLU        &    $32\times32\times256$      \\ \hline
Conv&     Conv4x4,stride=2        &    $16\times16\times512$      \\ 
&     LeakyReLU        &    $16\times16\times512$      \\ \hline
Conv&     Conv3x3,stride=1        &    $16\times16\times512$      \\ 
&     LeakyReLU        &    $16\times16\times512$      \\ \hline
Fully connected&     Reshape       &    $16\cdot16\cdot512$      \\ 
&     Linear        &    $1$      \\ \hline
\hline
\end{tabular}
\end{table}


\begin{table}[]
\centering
\small
\caption{\label{tab:resnet_g_small}ResNet Generator with $32\times32\times3$ resolution. Each ResNet block has a skip-connection that uses upsampling of its input and a 1x1 convolution. sBN denotes BN with self-modulation as proposed.}
\begin{tabular}{lcc}
\hline
\hline
Layer           & Details & Output size \\ \hline
Latent noise &  $\bm z\sim \mathcal{N}(0, I)$& $128$            \\\hline
Fully connected &  Linear& $4\cdot4\cdot256$            \\
         &  Reshape &$4\times4\times256$           \\ \hline
ResNet block   &    sBN, ReLU  &     $4\times4\times256$       \\
&Upsample&$8\times8\times256$\\
& Conv3x3, sBN, ReLU&$8\times8\times256$ \\
& Conv3x3 &$8\times8\times256$\\\hline
ResNet block   &    sBN, ReLU&     $8\times8\times256$         \\
&Upsample&$16\times16\times256$\\
& Conv3x3, sBN, ReLU&$16\times16\times256$ \\
& Conv3x3 &$16\times16\times256$\\\hline
ResNet block   &    sBN, ReLU&     $16\times16\times256$         \\
&Upsample&$32\times32\times256$\\
& Conv3x3, sBN, ReLU&$32\times32\times256$ \\
& Conv3x3 &$32\times32\times256$\\\hline
Conv& sBN, ReLU &$128\times128\times3$ \\
& Conv3x3, Tanh &$128\times128\times3$ \\\hline
\hline
\end{tabular}
\end{table}

\begin{table}[]
\centering
\small
\caption{\label{tab:resnet_d_small}ResNet Discriminator with $32\times32\times3$ resolution. Each ResNet block has a skip-connection that applies a 1x1 convolution with possible downsampling according to spatial dimension.}
\begin{tabular}{lcc}
\hline
\hline
Layer           & Details & Output size \\ \hline
Input image &  $ $& $32\times32\times3$            \\\hline
ResNet block   &   Conv3x3 &     $32\times32\times128$       \\
& ReLU,Conv3x3&$32\times32\times128$\\
& Downsample&$16\times16\times128$ \\\hline
ResNet block   &   ReLU,Conv3x3 &     $16\times16\times128$       \\
& ReLU,Conv3x3&$16\times16\times128$\\
& Downsample&$8\times8\times128$ \\\hline
ResNet block   &   ReLU,Conv3x3 &     $8\times8\times128$       \\
& ReLU,Conv3x3&$8\times8\times128$\\\hline
ResNet block   &   ReLU,Conv3x3 &     $8\times8\times128$       \\
& ReLU,Conv3x3&$8\times8\times128$\\\hline
Fully connected& ReLU,GlobalSum pooling&$128$\\\hline
& Linear&$1$\\\hline
\hline
\end{tabular}
\end{table}

\begin{table}[]
\centering
\small
\caption{\label{tab:resnet_g_large}ResNet Generator with $128\times128\times3$ resolution. Each ResNet block has a skip-connection that uses upsampling of its input and a 1x1 convolution. sBN denotes BN with self-modulation as proposed.}
\begin{tabular}{lcc}
\hline
\hline
Layer           & Details & Output size \\ \hline
Latent noise &  $\bm z\sim \mathcal{N}(0, I)$& $128$            \\\hline
Fully connected &  Linear& $4\cdot4\cdot1024$            \\
         &  Reshape &$4\times4\times1024$           \\ \hline
ResNet block   &    sBN, ReLU  &     $4\times4\times1024$       \\
&Upsample&$8\times8\times1024$\\
& Conv3x3, sBN, ReLU&$8\times8\times1024$ \\
& Conv3x3 &$8\times8\times1024$\\\hline
ResNet block   &    sBN, ReLU&     $8\times8\times1024$         \\
&Upsample&$16\times16\times1024$\\
& Conv3x3, sBN, ReLU&$16\times16\times1024$ \\
& Conv3x3 &$16\times16\times512$\\\hline
ResNet block   &    sBN, ReLU&     $16\times16\times512$         \\
&Upsample&$32\times32\times512$\\
& Conv3x3, sBN, ReLU&$32\times32\times512$ \\
& Conv3x3 &$32\times32\times256$\\\hline
ResNet block   &    sBN, ReLU&     $32\times32\times256$         \\
&Upsample&$64\times64\times256$\\
& Conv3x3, sBN, ReLU&$64\times64\times256$ \\
& Conv3x3 &$64\times64\times128$\\\hline
ResNet block   &    sBN, ReLU&     $64\times64\times128$         \\
&Upsample&$128\times128\times128$\\
& Conv3x3, sBN, ReLU&$128\times128\times128$ \\
& Conv3x3 &$128\times128\times64$\\\hline
Conv& sBN, ReLU &$128\times128\times3$ \\
& Conv3x3, Tanh &$128\times128\times3$ \\\hline
\hline
\end{tabular}
\end{table}

\begin{table}[]
\centering
\small
\caption{\label{tab:resnet_d_large}ResNet Discriminator with $128\times128\times3$ resolution. Each ResNet block has a skip-connection that applies a 1x1 convolution with possible downsampling according to spatial dimension.}
\begin{tabular}{lcc}
\hline
\hline
Layer           & Details & Output size \\ \hline
Input image &  $ $& $128\times128\times3$            \\\hline
ResNet block   &   Conv3x3 &     $128\times128\times64$       \\
& ReLU,Conv3x3&$128\times128\times64$\\
& Downsample&$64\times64\times64$ \\\hline
ResNet block   &   ReLU,Conv3x3 &     $64\times64\times64$       \\
& ReLU,Conv3x3&$64\times64\times128$\\
& Downsample&$32\times32\times128$ \\\hline
ResNet block   &   ReLU,Conv3x3 &     $32\times32\times128$       \\
& ReLU,Conv3x3&$32\times32\times256$\\
& Downsample&$16\times16\times256$ \\\hline
ResNet block   &   ReLU,Conv3x3 &     $16\times16\times256$       \\
& ReLU,Conv3x3&$16\times16\times512$\\
& Downsample&$8\times8\times512$ \\\hline
ResNet block   &   ReLU,Conv3x3 &     $8\times8\times512$       \\
& ReLU,Conv3x3&$8\times8\times1024$\\
& Downsample&$4\times4\times1024$ \\\hline
ResNet block   &   ReLU,Conv3x3 &     $4\times4\times1024$       \\
& ReLU,Conv3x3&$4\times4\times1024$\\\hline
Fully connected& ReLU,GlobalSum pooling&$1024$\\\hline
& Linear&$1$\\\hline
\hline
\end{tabular}
\end{table}

\end{document}